%% file: main.tex
\documentclass[10pt,twocolumn,letterpaper]{article}

\usepackage[table]{xcolor}
\usepackage[pagenumbers]{cvpr} 


\definecolor{cvprblue}{rgb}{0.21,0.49,0.74}
\usepackage[pagebackref,breaklinks,colorlinks,citecolor=cvprblue]{hyperref}
\usepackage{arydshln}
\usepackage{placeins}
\usepackage{subcaption} 
\usepackage{tabularx} 
\usepackage{arydshln}
\usepackage{booktabs}
\usepackage{multirow}
\usepackage{pifont} 

\usepackage{bbding} 
\def\paperID{14533} 
\def\confName{CVPR}
\def\confYear{2025}

\title{MEMO-Bench: A Multiple Benchmark for Text-to-Image and Multimodal Large Language Models on Human Emotion Analysis}

\author{Yingjie Zhou$^{1,3}$, Zicheng Zhang$^{1,3}$, Jiezhang Cao$^2$, Jun Jia$^{1,3}$, Yanwei Jiang$^{1,3}$, Farong Wen$^{1,3}$, \\Xiaohong Liu$^{1,3}$, Xiongkuo Min$^{1,3}$, Guangtao Zhai$^{1,3}$ \\
$^1$Shanghai Jiaotong University, $^2$Harvard Medical School, $^3$PengCheng Laboratory
}

\begin{document}

\twocolumn[{%
\renewcommand\twocolumn{1}{}{}%
\maketitle
\begin{center}
    \centering
    \vspace{-0.01cm}
    \includegraphics[width=0.91\textwidth]{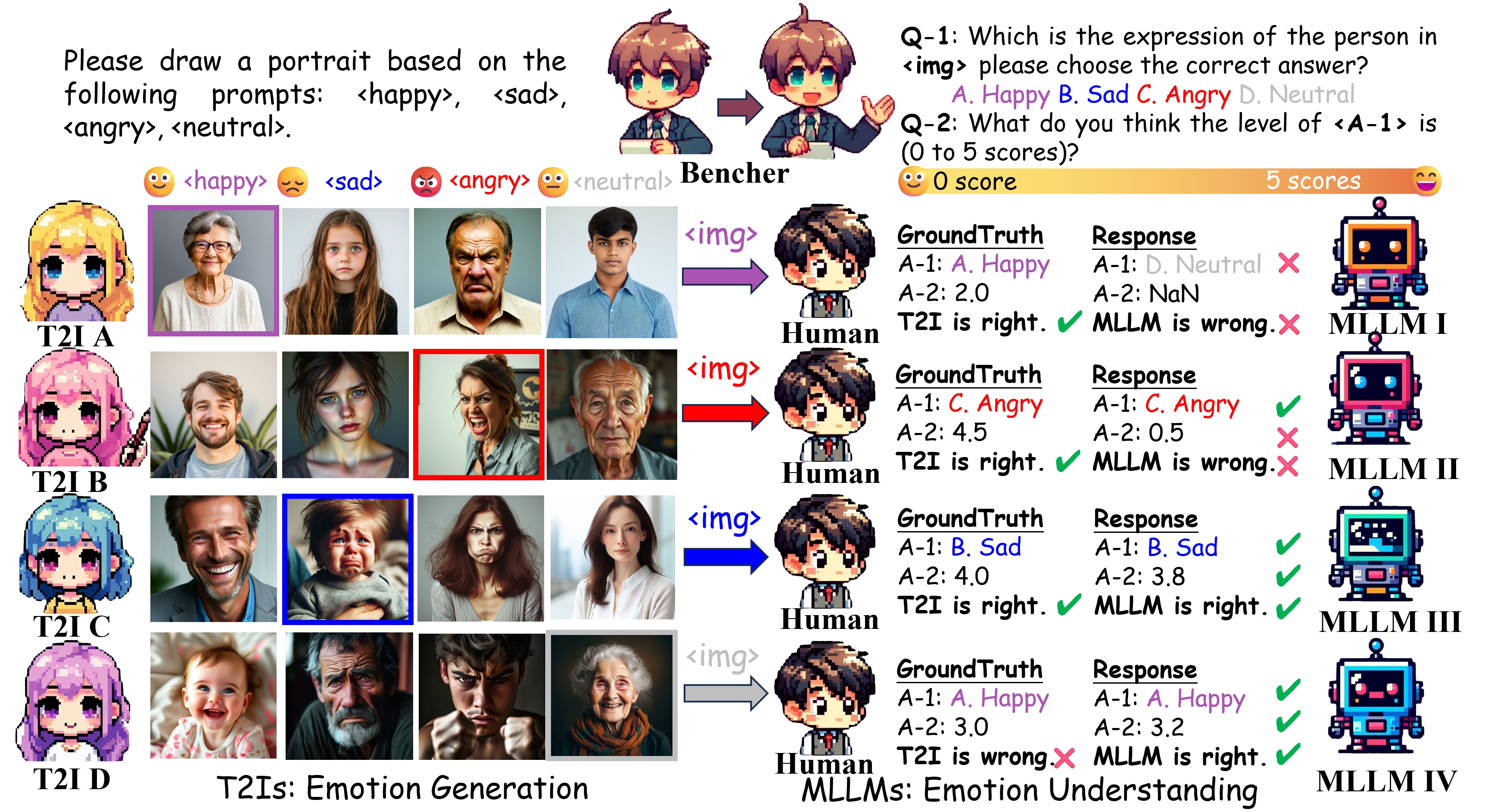}
    \captionof{figure}{MEMO-Bench's overall idea of human sentiment analysis review for AI. Specifically, it includes the evaluation of emotion generation ability for Text-to-Image models and the evaluation of emotion comprehension ability for multimodal language large models.}
\end{center}
}]

\input{sec/0_abstract}    
\input{sec/1_intro}

\input{sec/2_formatting}

\input{sec/3_finalcopy}

\input{sec/4_bench1}

\input{sec/5_bench2}
\input{sec/6_con}
{
    \small
    \bibliographystyle{ieeenat_fullname}
    \bibliography{main}
}


\end{document}

%% file: sec/0_abstract.tex
\begin{abstract}
Artificial Intelligence (AI) has demonstrated significant capabilities in various fields, and in areas such as human-computer interaction (HCI), embodied intelligence, and the design and animation of virtual digital humans, both practitioners and users are increasingly concerned with AI’s ability to understand and express emotion. Consequently, the question of whether AI can accurately interpret human emotions remains a critical challenge. To date, two primary classes of AI models have been involved in human emotion analysis: generative models and Multimodal Large Language Models (MLLMs). To assess the emotional capabilities of these two classes of models, this study introduces MEMO-Bench, a comprehensive benchmark consisting of 7,145 portraits, each depicting one of six different emotions, generated by 12 Text-to-Image (T2I) models. Unlike previous works, MEMO-Bench provides a framework for evaluating both T2I models and MLLMs in the context of sentiment analysis. Additionally, a progressive evaluation approach is employed, moving from coarse-grained to fine-grained metrics, to offer a more detailed and comprehensive assessment of the sentiment analysis capabilities of MLLMs. The experimental results demonstrate that existing T2I models are more effective at generating positive emotions than negative ones. Meanwhile, although MLLMs show a certain degree of effectiveness in distinguishing and recognizing human emotions, they fall short of human-level accuracy, particularly in fine-grained emotion analysis. The MEMO-Bench will be made publicly available to support further research in this area.
\end{abstract}

%% file: sec/1_intro.tex
\section{Introduction}
\label{sec:intro}
Currently, artificial intelligence (AI) has achieved capabilities comparable to, and in some cases surpassing, human performance across a variety of domains, suggesting that it possesses a certain degree of rational thinking. However, the presence of rationality alone does not equate to complete ``intelligence," as the question of whether AI truly experiences emotions remains unresolved \cite{grassini2024understanding,cyran2024new,assunccao2022overview}. In real-world applications, such as human-computer interaction (HCI) \cite{barricelli2024digital,wadley2022future,mikuckas2014emotion} and embodied intelligence \cite{ren2024embodied,duan2022survey}, as well as in the design of immersive media like virtual digital humans \cite{cumt,yang2024emogen,eat,zhou2021omnidirectional}, user emotions play a critical role in shaping AI decision-making processes. Unfortunately, most existing interactive systems rely primarily on text-based interfaces \cite{phongphaew2024text}, rather than leveraging multimodal information, such as visual and auditory cues, to effectively capture and understand emotional fluctuations. This limitation significantly restricts the emotional comprehension capabilities of these systems. The advent of Multimodal Large Language Models (MLLMs) introduces new possibilities for emotion-aware interactions, enabling AI to potentially understand human emotions. However, the extent of this ability remains largely unexplored. Existing research \cite{lian2024merbench} has focused predominantly on assessing the capacity of MLLMs to recognize and categorize emotions, often overlooking their ability to understand emotions at a more granular level. Moreover, the requirement for emotion understanding extends beyond MLLMs to include Text-to-Image (T2I) models, which are increasingly used to generate character portraits that convey specific emotional states. Consequently, evaluating AI's emotion analysis capabilities should encompass both MLLMs and T2I models, as both are integral to advancing emotion-aware AI systems.

In this paper, we introduce a \underline{m}ultiple \underline{emo}tion analysis \underline{bench}mark, MEMO-Bench, designed to evaluate the emotional analysis capabilities of both T2I models and MLLMs. MEMO-Bench consists of 7,145 AI-generated portrait images (AGPIs), representing six distinct emotional states. Specifically, for each emotion, we employ 100 prompts, which are provided to T2I models for portrait generation. Upon generation, the resulting AGPIs are manually reviewed to assess the alignment between the images and the intended emotional expression in prompts, allowing us to evaluate the emotion generation abilities of T2I models. For MLLM evaluation, we employ a progressive emotion assessment approach that ranges from coarse-grained to fine-grained analysis. Initially, the MLLM is tasked with categorizing the emotions of the AGPIs. Subsequently, the MLLM is presented with the correctly categorized portraits and asked to assess their emotional intensity at a finer level.The experimental results show that the existing T2I model has some sentiment generation capability, but is still limited in negative sentiment generation. MLLM, on the other hand, performs relatively well in coarse-grained emotion categorization, but is completely incapable of accurately understanding the degree of fine-grained human emotions. These findings highlight the current limitations in AI’s ability to fully comprehend human emotions, providing valuable insights for the development of more advanced emotion-aware AI systems. The main contributions of this paper are as follows:


\begin{itemize}
\item A large-scale dataset, MEMO-Bench, is introduced to assess the sentiment analysis capabilities of AI. This dataset includes 7,145 emotionally generated AGPIs across six different emotions, produced by twelve T2I models.
\item Based on MEMO-Bench, a multiple benchmark framework for AI's sentiment analysis capability is proposed. The benchmark focuses on the sentiment generation capability of T2I models as well as tests the sentiment understanding capability of MLLMs.
\item An asymptotic approach is applied for emotion comprehension evaluation. MLLMs are queried with both coarse-grained emotion categorization tasks and fine-grained emotion level analysis tasks to provide a thorough assessment of their emotional understanding capabilities.
\end{itemize}

%% file: sec/2_formatting.tex
\begin{figure*}[!t]
    \centering
    \subfloat[HAP]{\includegraphics[width=0.333\linewidth]{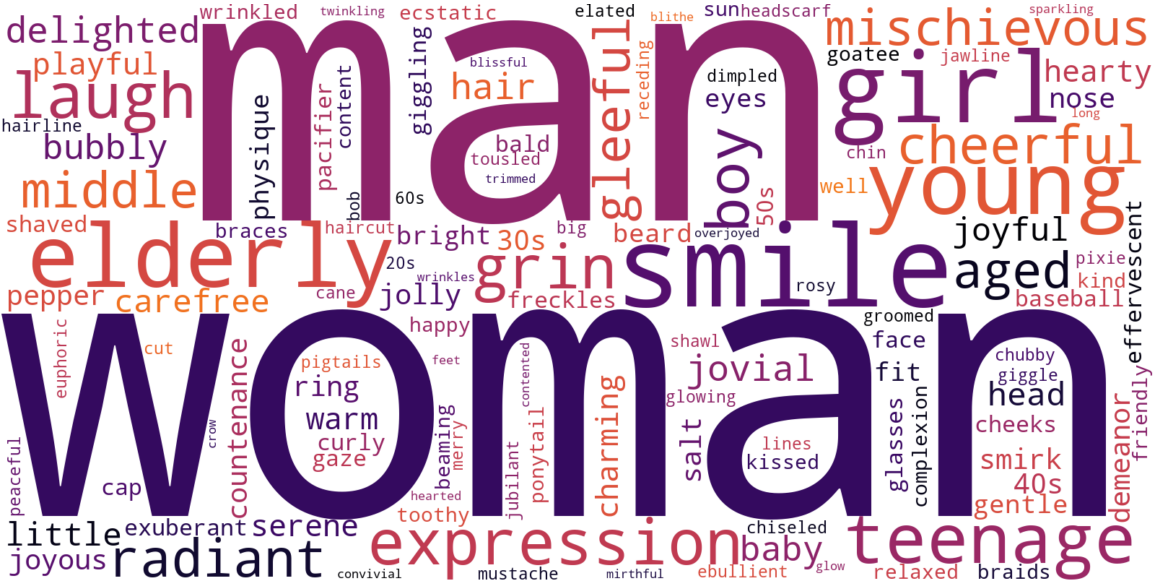}}
    \subfloat[WOR]{\includegraphics[width=0.333\linewidth]{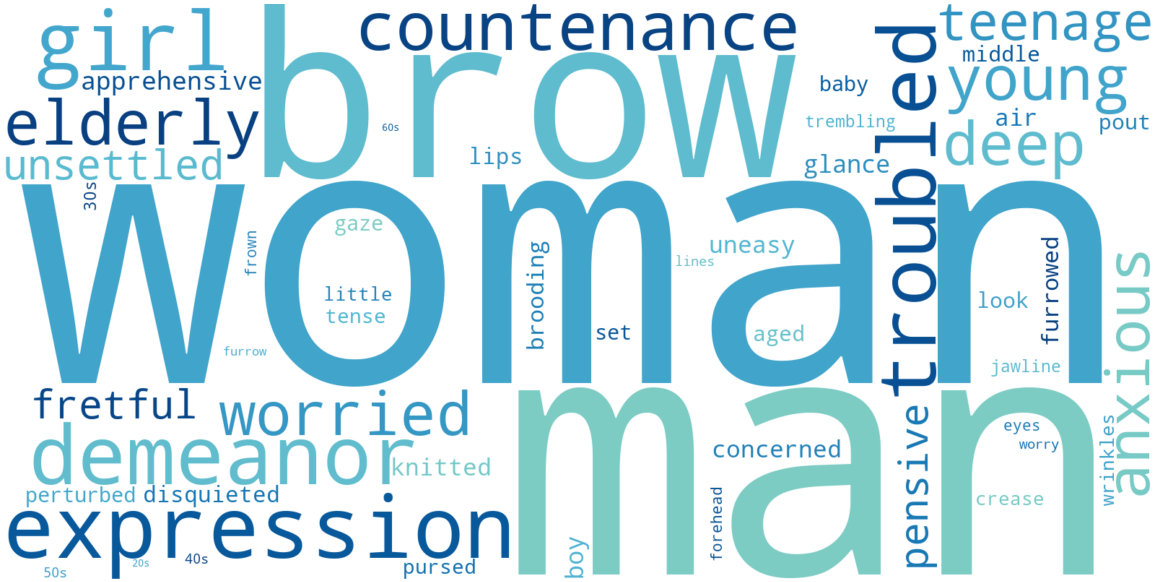}}
    \subfloat[SUR]{\includegraphics[width=0.333\linewidth]{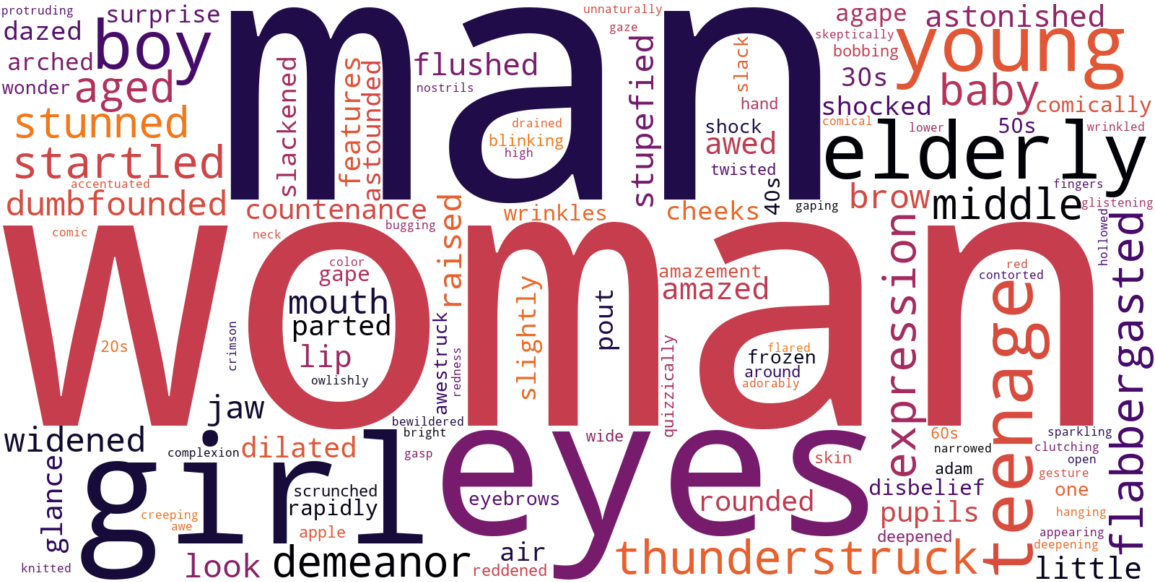}}
    
    \subfloat[SAD]{\includegraphics[width=0.333\linewidth]{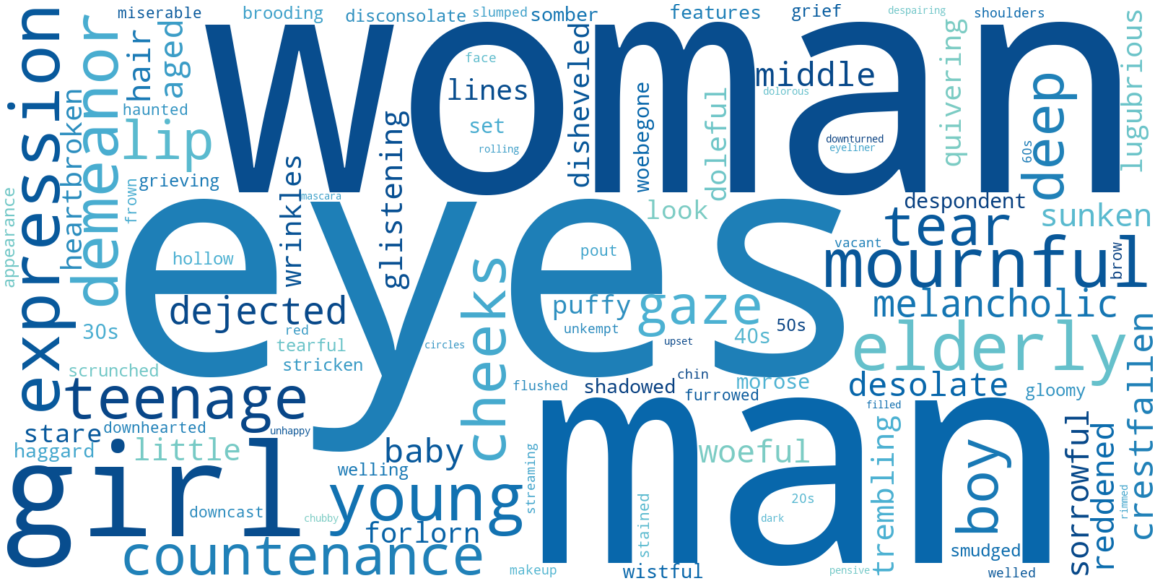}}
    \subfloat[NEU]{\includegraphics[width=0.333\linewidth]{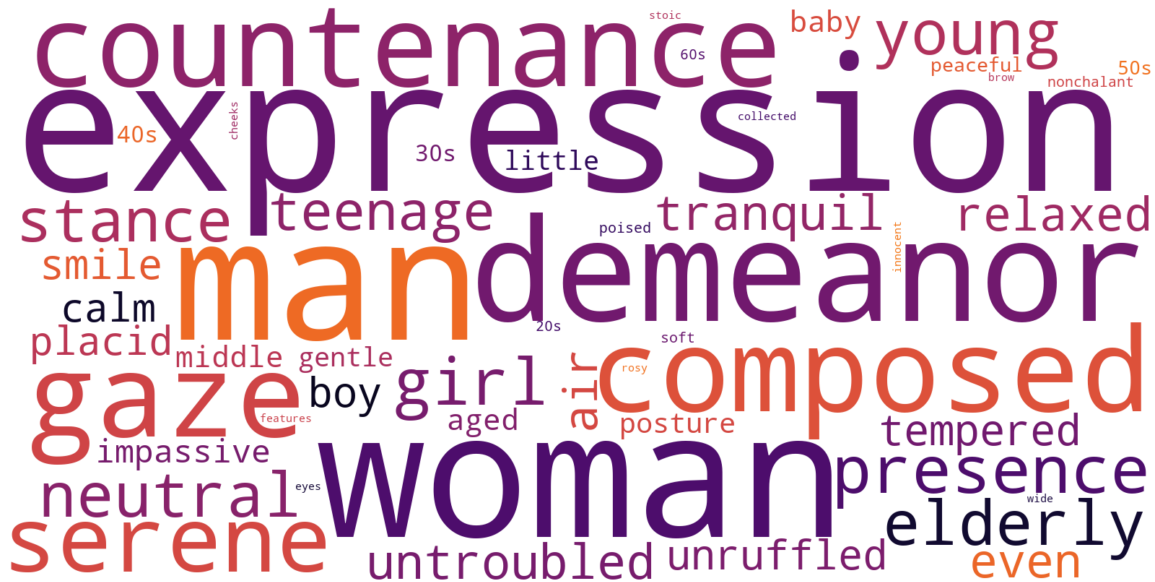}}
    \subfloat[ANG]{\includegraphics[width=0.333\linewidth]{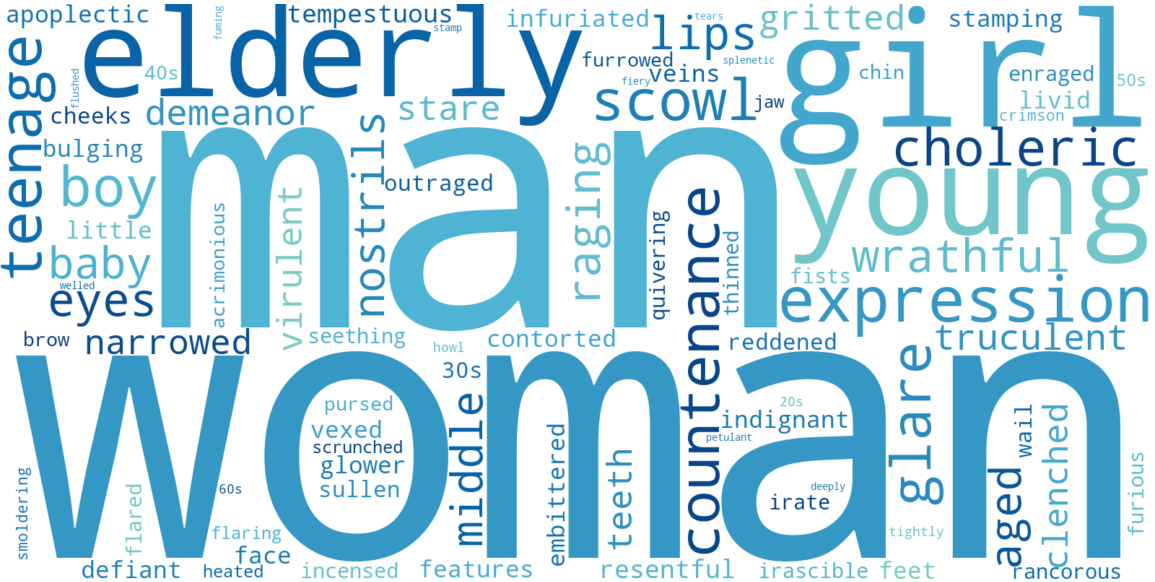}}
    
    
    \vspace{-0.4cm}
    \caption{Prompts for generating different emotions. Warm colors indicate positive emotions and cool colors indicate negative emotions.}
    \label{fig:prompts}
    \vspace{-0.4cm}
\end{figure*}

\section{Related Works}
\subsection{Affective Computing}
Affective computing, also known as Emotion AI \cite{mantello2024emotional,khare2024emotion}, is a key domain within AI that focuses on enabling machines to recognize, interpret, and simulate human emotions. With the emergence of technologies such as virtual digital humans, large language models (LLMs), and embodied intelligence, the integration of human-like emotions into AI has become an area of significant interest \cite{kim2024self,scaria2023instructabsa,stigall2024large,sun2023sentiment,wang2023enhance,wang2024wisdom,gan2023usa,fei2023reasoning}. Affective computing can be broadly categorized into two interrelated tasks: Affect Generation (AG) \cite{pei2024affective} and Affect Understanding (AU) \cite{zhao2023chatgpt}, both of which are critical for the continued advancement of AI systems \cite{zhang2024affective}. AG refers to the capability of AI to create content that conveys specific or contextually appropriate emotions, either based on human input or situational cues. In contrast, AU involves the ability of AI to accurately interpret the emotions present in diverse forms of multimodal data. Traditionally, sentiment analysis and affective computing rely on pre-trained language models \cite{devlin2018bert,liu2019roberta} that are trained on manually labeled datasets \cite{amin2023will,tan2023survey,krugmann2024sentiment}; however, these models often exhibit limited generalization ability due to constraints in both model architecture and dataset size. In contrast, LLMs have introduced a new paradigm for affective computation, leveraging extensive datasets and advanced model architectures to improve performance. Despite the progress made by LLMs in emotion-related tasks \cite{wei2022emergent,zhang2023sentiment,amin2024wide}, the question of whether AI is truly capable of achieving emotional awareness remains an open and critical area of research.

\subsection{Benchmarks for AI}
As the influence of AI, particularly LLMs, on human life continues to deepen, there has been a growing interest among scholars to quantify the performance of these models. This has led to the creation of reliable benchmarks that can guide future research and model development. In recent years, a variety of benchmarks have been established to evaluate the capabilities of LLMs across different domains. Prominent examples include C-Eval \cite{huang2024c}, AGI-Eval \cite{zhong2023agieval}, MMLU \cite{hendrycks2020measuring}, and CMMLU \cite{li2023cmmlu}, which assess the proficiency of LLMs in various academic and practical disciplines. Besides, in the field of computer vision, benchmarks such as Q-Bench \cite{zhang2024q,zhang2024qv} and A-Bench \cite{zhang2024bench} have been developed to evaluate LLMs' ability to perceive the quality of digital media \cite{zhang2024quality}. Although there have been benchmarks \cite{lian2024merbench,sabour2024emobench} designed to evaluate the emotion-perception abilities of LLMs, most existing frameworks still rely on coarse-grained assessments that fail to account for the nuanced levels of emotional understanding. Furthermore, existing work primarily focuses on the emotional comprehension abilities of LLMs, often neglecting the emotional capabilities of generative models. In response to this gap, the MEMO-Bench introduced in this paper aims to provide a more comprehensive evaluation by considering three key aspects, namely, the emotional generation capabilities of generative models, the quality of content generated by these models, and the emotional comprehension abilities of MLLMs. This holistic approach offers a more integrated and thorough assessment of AI's capacity for emotion analysis.

\begin{figure*}[!t]
    
    \centering
    \includegraphics[width =1\linewidth]{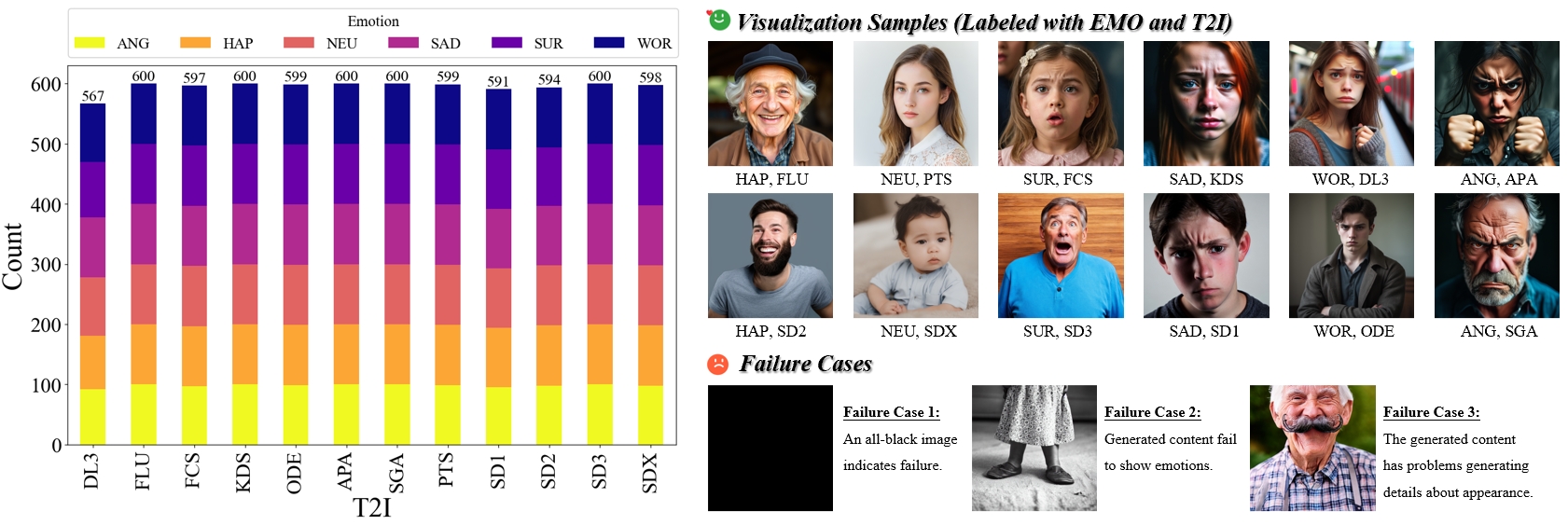}
    \vspace{-0.7cm}
    \caption{Visualization of various T2I models' performance. On the left is the number of AGPIs that can be successfully generated by each type of T2I model based on different sentiment prompts, and on the right is a sample of AGPIs generated by each type of T2I model, including both successful and unsuccessful cases.}
    \label{fig:disvis}
    \vspace{-0.5cm}
\end{figure*}

%% file: sec/3_finalcopy.tex
\begin{table}[!tp]
    \centering
    \caption{Details of T2I models employed for the generation. }
    \vspace{-0cm}
    \resizebox{1\linewidth}{!}{\begin{tabular}{c|c|c|c|c}
    \toprule
         Type & Label &T2I Model & Year  & Output Resolution\\ \hline
          \multirow{1}{*}{Closed source} &DL3 &Dalle3 \cite{betker2023improving} &2023 &1,024×1,024 \\     \hdashline
          \multirow{11}{*}{Open source} &SD2&Stable Diffusion 2.1 \cite{Rombach_2022_CVPR}&2022  &512×512 \\
          &SD1&Stable Diffusion 1.5 \cite{Rombach_2022_CVPR}&2022  &512×512  \\
          &SDX&Stable Diffusion XL \cite{podell2023sdxl}&2023  &1,024×1,024  \\
         &FCS&Fooocus \cite{fooocus}&2023  &1,024×1,024 \\
         &KDS&Kandinsky \cite{kandinsky}&2023 &1,024×1,024 \\
         &ODE&OpenDalleV1.1 \cite{opendalle}&2023  &1,024×1,024 \\
         &PTS&Proteus \cite{proteus}&2023  &1,024×1,024 \\ 
         &APA&PixArt-$\alpha$ \cite{chen2023pixartalpha}&2023  &1,024×1,024 \\ 
         &SGA&PixArt-$\Sigma$ \cite{chen2024pixartdelta}&2024  &1,024×1,024 \\ 
         &SD3&Stable Diffusion 3 \cite{esser2024scaling}&2024  &1,024×1,024 \\
         &FLU&FLUX.1 \cite{flux}&2024  &1,024×1,024 \\
         
    \bottomrule
    \end{tabular}}
    \label{tab:t2i}
    \vspace{-0.2cm}
\end{table}
\section{Construction of MEMO-Bench}
\subsection{Emotions and Prompts}
Building on prior research \cite{lian2024merbench}, six fundamental emotions are selected to construct the MEMO-Bench dataset: happiness (HAP), sadness (SAD), anger (ANG), surprise (SUR), worry (WOR), and neutrality (NEU). For each emotion, 100 distinct prompts are designed to generate AGPIs that convey the respective emotion. To provide a clearer visualization of the selected prompts, Fig.~\ref{fig:disvis}, most prompts consist primarily of emotion-related words and phrases, with the direct inclusion of specific emotional terms such as \textit{``Happy"} and \textit{``Sad."} This highlights the additional challenge posed to T2I models in generating emotionally expressive content based on more nuanced, context-driven prompts.

\subsection{Text-to-Image Models}
To comprehensively evaluate the sentiment generation capabilities of existing T2I models and to provide suitable visual data for assessing the sentiment comprehension abilities of MLLMs, we select 12 representative T2I models. The specific details of these models are outlined in Table~\ref{tab:t2i}. Additionally, Fig.~\ref{fig:disvis} presents some typical cases that illustrate the variation in performance across different models. As shown in the figure, even with identical prompts, the images generated by different T2I models exhibit significant differences in both image quality and emotional expression. In some instances, certain models produce images that are either unrelated to the prompts or fail to convey any emotion. After manually reviewing all the generated images, we retain a total of 7,145 emotionally expressive images and recorded the number of selected images for each T2I model. It is also important to note that for the closed-source DL3 model, certain \textit{``Angry"} prompts trigger sensitivity issues, resulting in a high number of generation failures.


    

\subsection{Subjective Annotation}
\label{sec:sub}
In contrast to existing studies that primarily rely on GPT-4o responses for annotation \cite{themselvesautobench}, this paper ensures the reliability and validity of the annotation process by subjectively annotating the 7,145 AGPIs. This is achieved through the recruitment of volunteers who assess the images across three dimensions: sentiment category, sentiment intensity, and image quality. Specifically, 15 male and 14 female participants are invited to participate in the subjective annotation, following the guidelines outlined in ITU-R B.T. 500-13 \cite{bt2002methodology}. The annotation process takes place in a well-controlled laboratory environment to maintain consistency. To facilitate the presentation and annotation of the generated images, we design a subjective annotation platform using Gradio \cite{abid2019gradio}. The platform includes modules for image quality assessment, sentiment recognition and level analysis and is displayed on an iMac monitor with a resolution of 4096×2304. The entire subjective annotation process is divided into 15 phases, with no more than 500 AGPI annotation tasks per phase. To ensure the quality and reliability of the annotations, volunteers are instructed to take a break of at least 15 minutes between each phase, with a maximum of four phases completed per day. Prior to the start of the first phase, all volunteers underwent a 30-minute training session, which included an introduction to the annotation task and the platform interface.


At the end of the annotation process, a total of 207,205 annotation sets are collected. Each annotation for the $j$-th AGPI by the $i$-th volunteer can be represented as a ternary tuple ($e_t$, $e_d$, $q$), where $e_t$ denotes the emotion category, $e_d$ refers to the degree of emotion, and $q$ represents image quality. For sentiment categorization, the emotion $e_t$ with the highest frequency of occurrence is selected as the final sentiment category $E_T$ of the generated image. In the case of sentiment intensity and image quality, z-scores \cite{zhang2023subjective,thqa3d,3dgcqa,reliqa,dhhqa,ddhqa,h3d,6gqa} are applied to standardize values of $e_d$ and $q$:

\begin{equation}
z_{ij}^u = \frac{{{u_{ij}} - \mu _i^u}}{{\sigma _i^u}},u \in [{e_d},q],
\end{equation}
where $\mu _i^u = \frac{1}{{{N_i}}}\sum\limits_{j = 1}^{{N_i}} {{u_{ij}}} $, $\sigma _i^u = \sqrt {\frac{1}{{{N_i} - 1}}\sum\limits_{j = 1}^{{N_i}} {\left( {{u_{ij}} - {\mu _i}} \right)} } $, and $N_i$ denotes the total number of AGPIs evaluated by the $i$-th subject. In accordance with the rejection procedure outlined in \cite{bt2002methodology}, ratings from unreliable subjects are excluded. The remaining z-scores $z_{ij}$ are linearly rescaled to range [0, 5]. Finally, the mean opinion scores (MOSs) for the $j$-th AGPI are computed by averaging the rescaled z-scores:
\begin{equation}
M O S^u_{j}=\frac{1}{M} \sum_{i=1}^{M} z_{i j}^{u'}, u \in [{e_d},q],
\end{equation}
where $M$ denotes the number of the valid subjects, and $z_{i j}^{'}$ represents the rescaled z-scores.

%% file: sec/4_bench1.tex
\section{Benchmark for T2I Models}
\subsection{Experiment Setup}
The evaluation of the emotional generation capability of T2I models encompasses two critical dimensions: the quality of the AGPIs and the accuracy of emotion synthesis. To assess these aspects, a comprehensive evaluation framework is employed, which integrates subjective annotations and customized prompts. This approach allows for the assessment of both the visual quality and the emotional fidelity of AGPIs, providing a means to quantify the emotional generation proficiency of various T2I models. Image quality is evaluated using the Mean Opinion Scores (MOSs) of AGPIs as a benchmark indicator. In contrast, the accuracy of emotion generation, denoted as $GACC_{k}$ for the $k$th emotion, can be initially formulated as follows:

\begin{figure}[!t]
    
    \centering
    \includegraphics[width =0.95\linewidth]{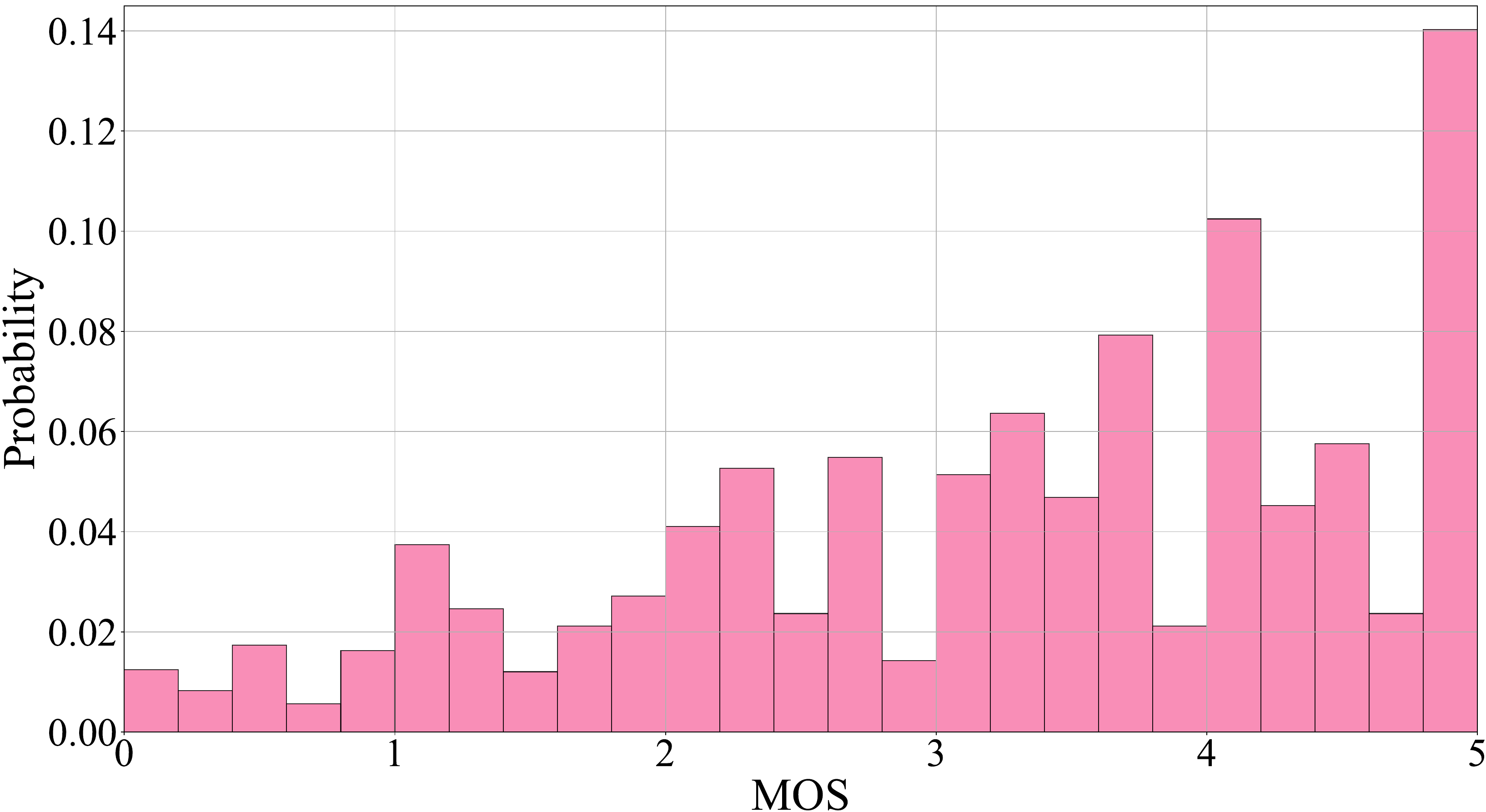}
    \vspace{-0.3cm}
    \caption{Distribution of MOSs for all AGPIs.}
    \label{fig:mosdis}
    \vspace{-0.3cm}
\end{figure}

\begin{figure}[!t]
    \centering
    
    \subfloat[]{\includegraphics[width=0.5\linewidth]{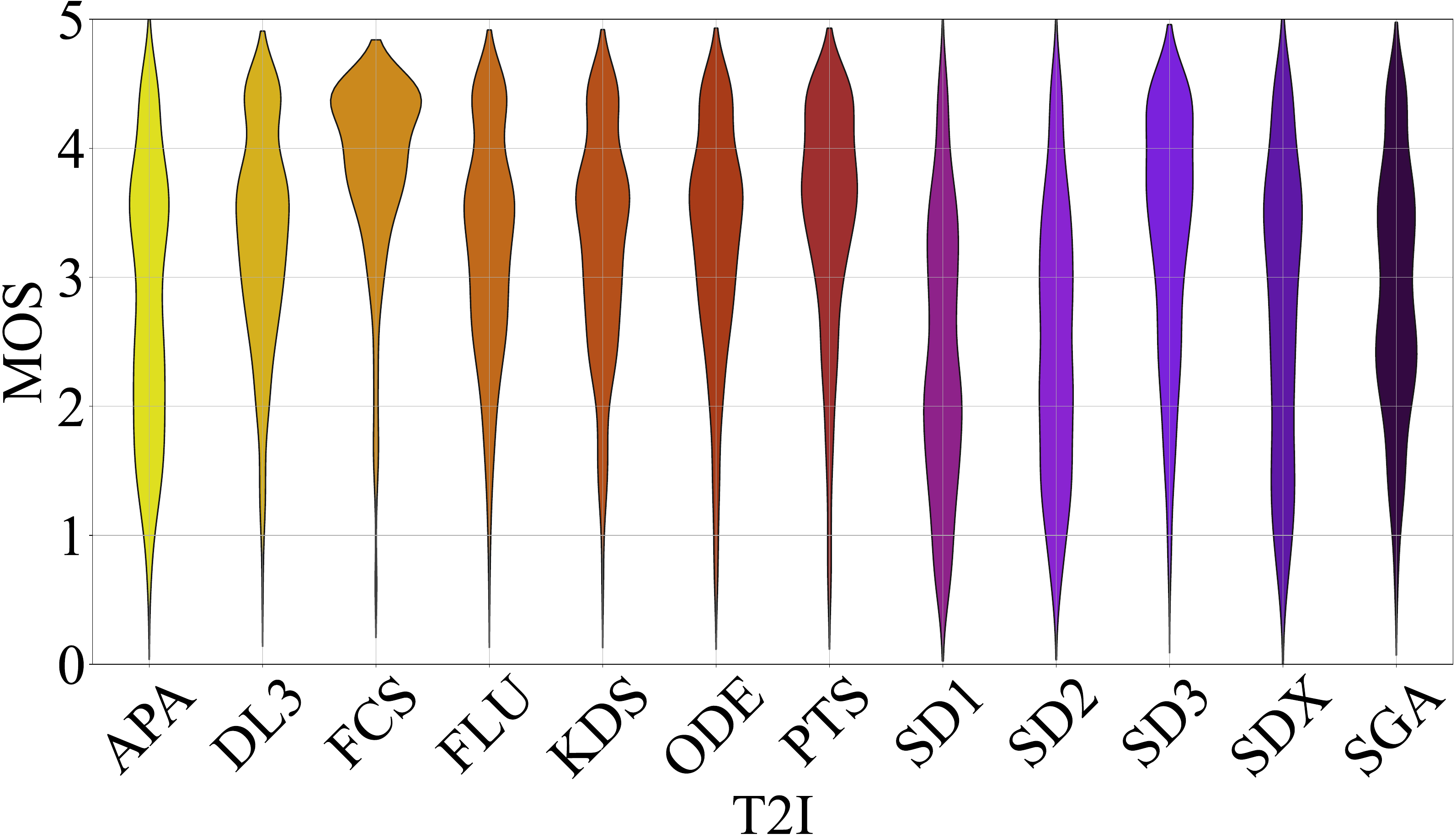}}
    \subfloat[]{\includegraphics[width=0.492\linewidth]{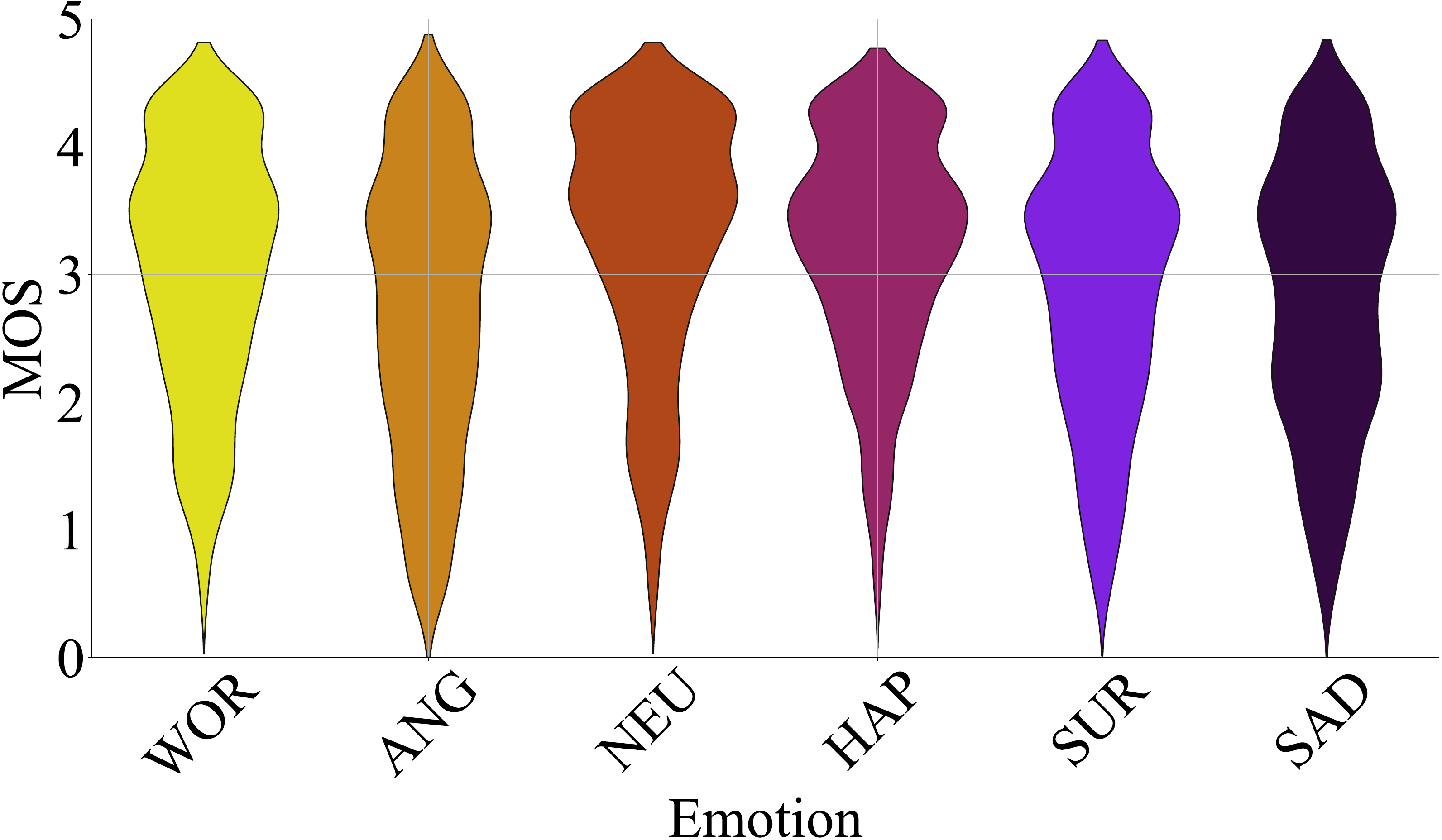}}
    
    
    \vspace{-0.3cm}
    \caption{Effect of different factors on the distribution of MOSs.}
    \label{fig:mosviolin}
    \vspace{-0.6cm}
\end{figure}

\begin{equation}
\label{lab:gacc}
GAC{C_k} = P(E_T^k|G{P_k}),k = 1,2, \ldots ,6,
\end{equation}
where $G{P_k}$ represents the complete set of prompts employed for generating the $k$th emotion, excluding instances of T2I model generation failures. Furthermore, $E_{T}^{k}$ signifies the full set of AGPIs labeled with the $k$th emotion. Additionally, to further investigate the different types of errors, the occurrences of generation errors across various emotion prompts are captured using the generation error rate, denoted as $GERR_{kl}$:

\begin{equation}
\label{lab:gerr}
GER{R_{kl}} = P(E_T^l|G{P_k}),k,l = 1,2, \ldots ,6,k\neq l.
\end{equation}
The $GACC$ and $GERR$ serve as essential tools for evaluating the T2I model's ability to accurately synthesize human emotions, which provide a comprehensive assessment of both the precision and error rates in emotion generation.

\begin{table}[!t]
\centering
\renewcommand\tabcolsep{1.5pt} 
\caption{Performance of T2Is. Best in {\bf\textcolor{red}{RED}}, second in \bf\textcolor{blue}{BLUE}.}
\begin{tabular}{c|ccccccc}
\toprule
\multirow{2}{*}{T2I}  & \multicolumn{7}{c}{$GACC$}  \\ 
\cline{2-8}
 & HAP& SAD& WOR       & NEU      & SUR     & ANG     & $All$   \\ \hline
\rowcolor{gray!25}APA & \bf\textcolor{red}{1.0000} & \bf\textcolor{blue}{0.5900} & 0.3600 & 0.6600 & \bf\textcolor{blue}{0.8300} & 0.7400 & 0.6967             \\
DL3 & 0.9683 & 0.5500 & \bf\textcolor{blue}{0.3807} & 0.7005 & 0.6667 & 0.7120 & \bf\textcolor{blue}{0.7160}             \\
\rowcolor{gray!25}FCS & 0.9000 & 0.1200 & 0.0700 & \bf\textcolor{red}{0.8000} & 0.2800 & 0.1546 & 0.3886             \\
FLU & \bf\textcolor{blue}{0.9900} & 0.3300 & 0.3400 & 0.6700 & 0.6000 & \bf\textcolor{blue}{0.8100} &  0.6233            \\
\rowcolor{gray!25}KDS & 0.9600 & 0.1600 & 0.1300 &  0.6100 & 0.5200 & 0.2600 &  0.4400            \\
ODE & 0.9600 & 0.4600 & 0.3100 & 0.6800 & 0.5600 & 0.6768 &  0.6077            \\
\rowcolor{gray!25}PTS & 0.9300 & 0.3500 & 0.2600 & \bf\textcolor{red}{0.8000} & 0.5400 & 0.3838 & 0.5442             \\
SD1 & 0.9596 & 0.5556 & 0.2400 & 0.6162 & 0.1919 & 0.3368 &  0.4839            \\
\rowcolor{gray!25}SD2 & 0.9500 & 0.3838 & 0.3200 & 0.5600 & 0.3918 & 0.1224 & 0.4562             \\
SDX & 0.9800 & 0.3700 & 0.2200 & 0.7000 & 0.4200 & 0.2959 & 0.4983             \\
\rowcolor{gray!25}SD3 & 0.9800 & 0.2300 & 0.1200 & 0.6600 & 0.3300 & 0.3000 &   0.4367          \\
SGA & 0.9000 & \bf\textcolor{red}{0.6000} & \bf\textcolor{red}{0.4200} & \bf\textcolor{blue}{0.7600} & \bf\textcolor{red}{0.8700} & \bf\textcolor{red}{0.9200} &   \bf\textcolor{red}{0.7450}           \\

\bottomrule
\end{tabular}
\label{tab:t2iper}
\vspace{-0.6cm}
\end{table}

\begin{figure*}[!t]
    \centering
    \subfloat[APA]{\includegraphics[width=0.245\linewidth]{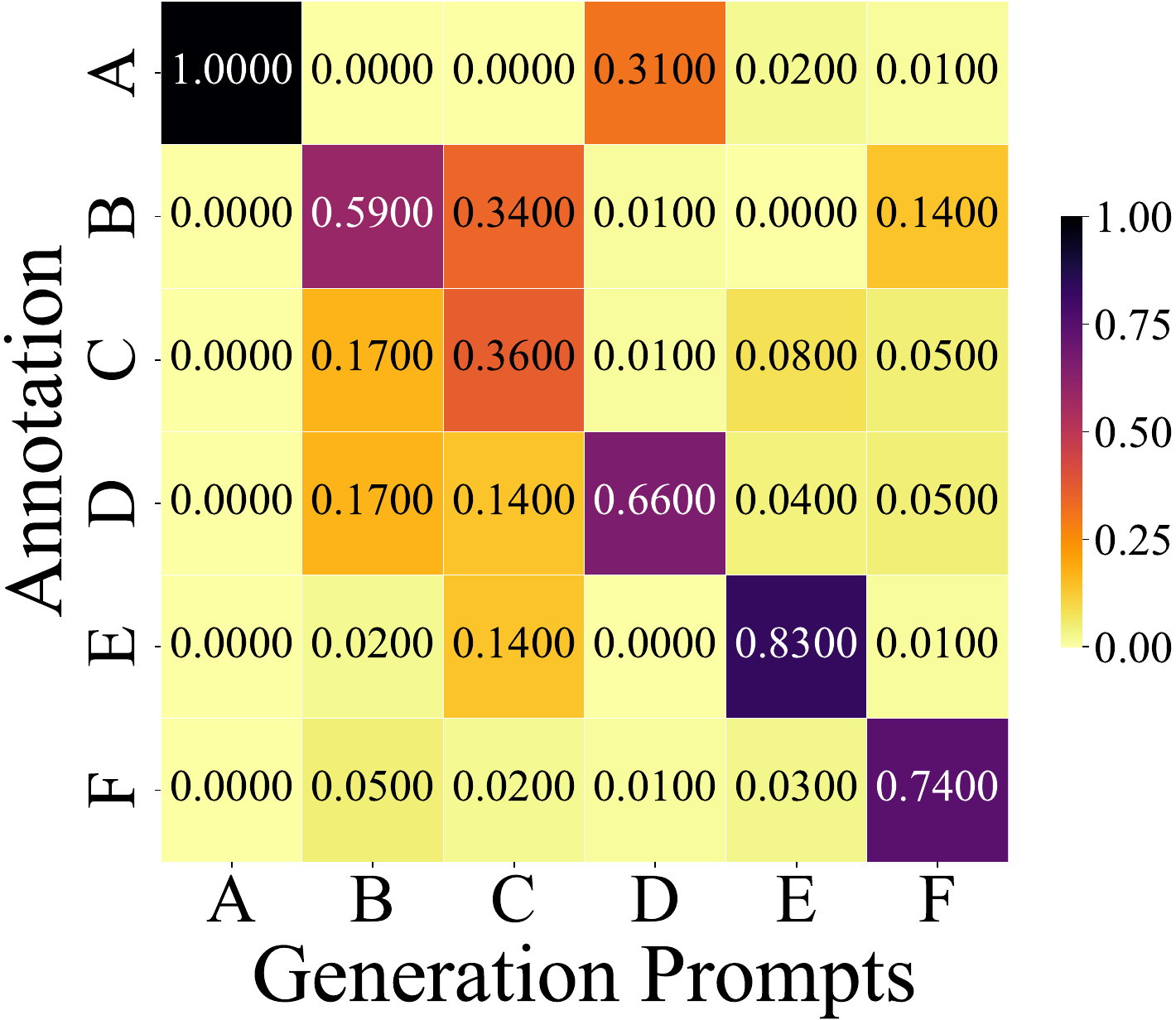}}
    \subfloat[DL3]{\includegraphics[width=0.245\linewidth]{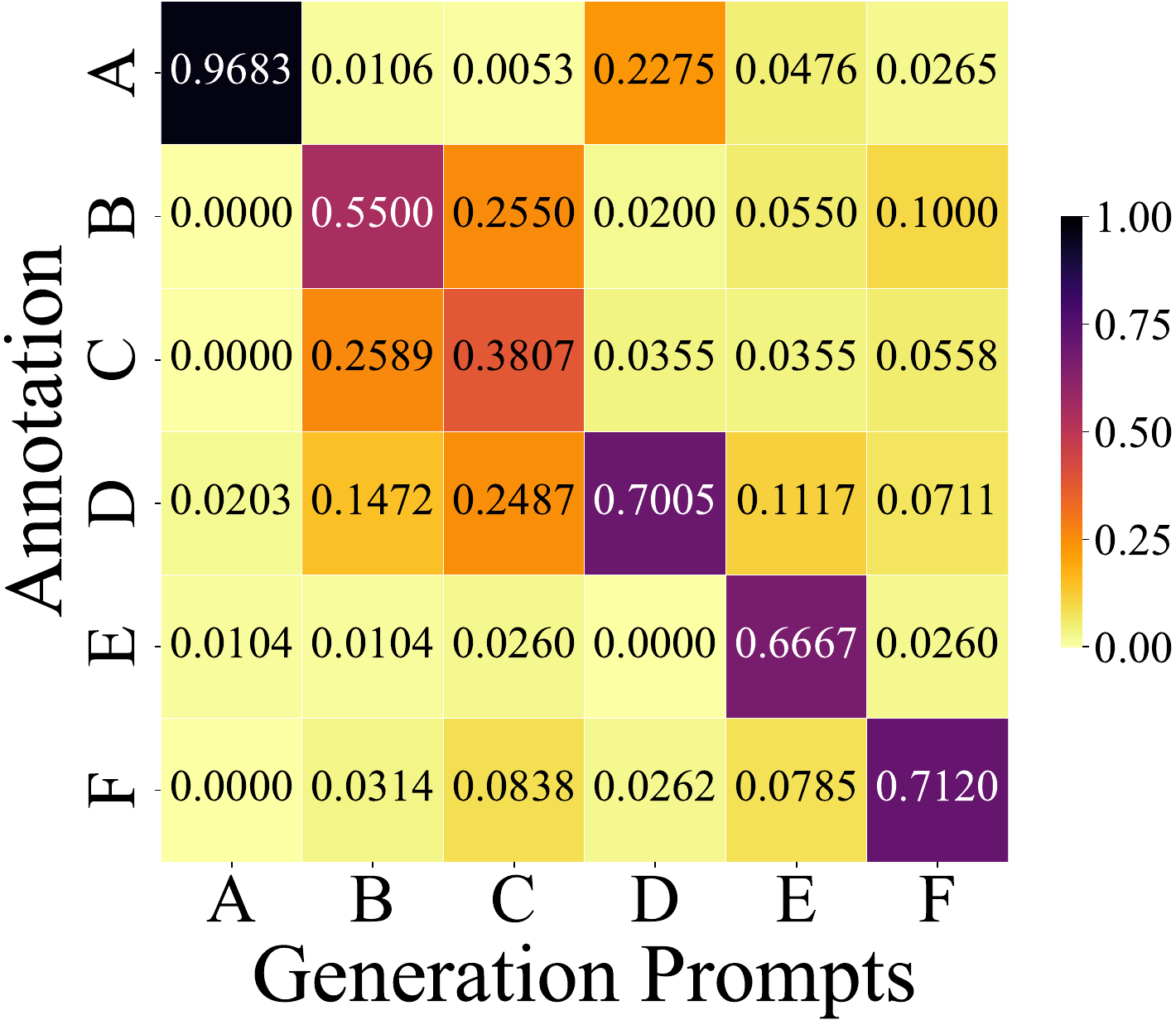}}
    \subfloat[FCS]{\includegraphics[width=0.245\linewidth]{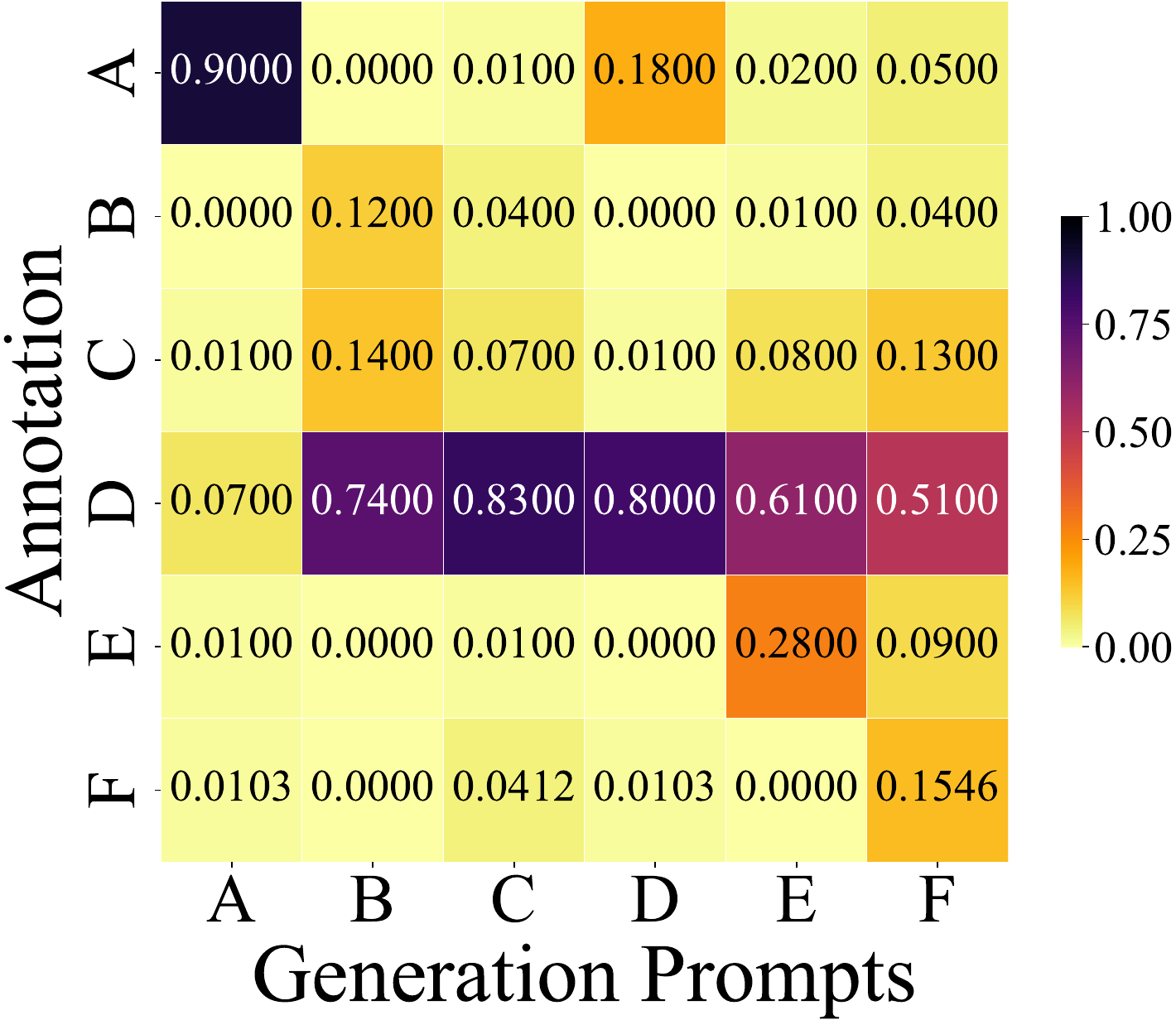}}
    \subfloat[FLU]{\includegraphics[width=0.245\linewidth]{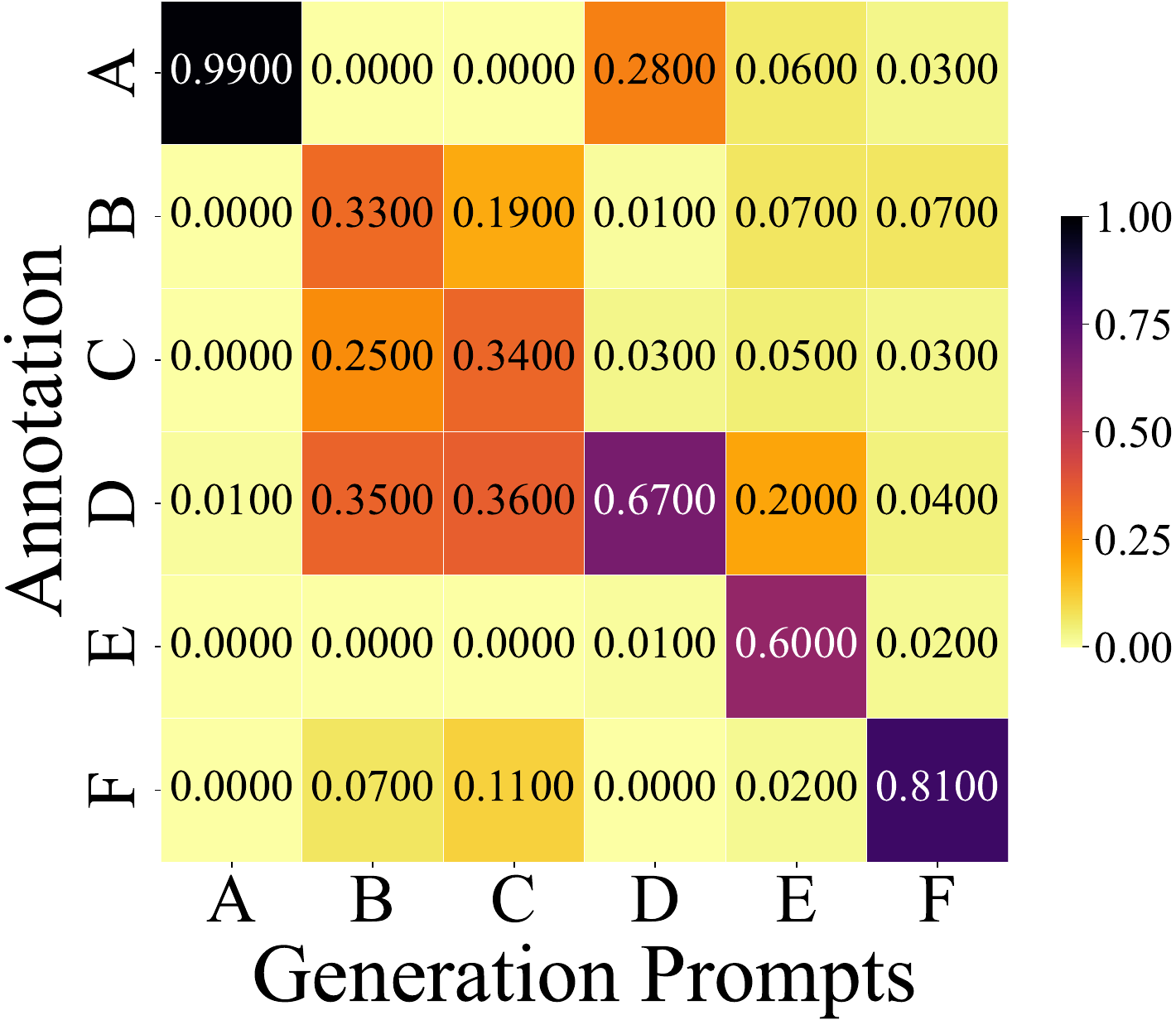}}
    
    \subfloat[KDS]{\includegraphics[width=0.245\linewidth]{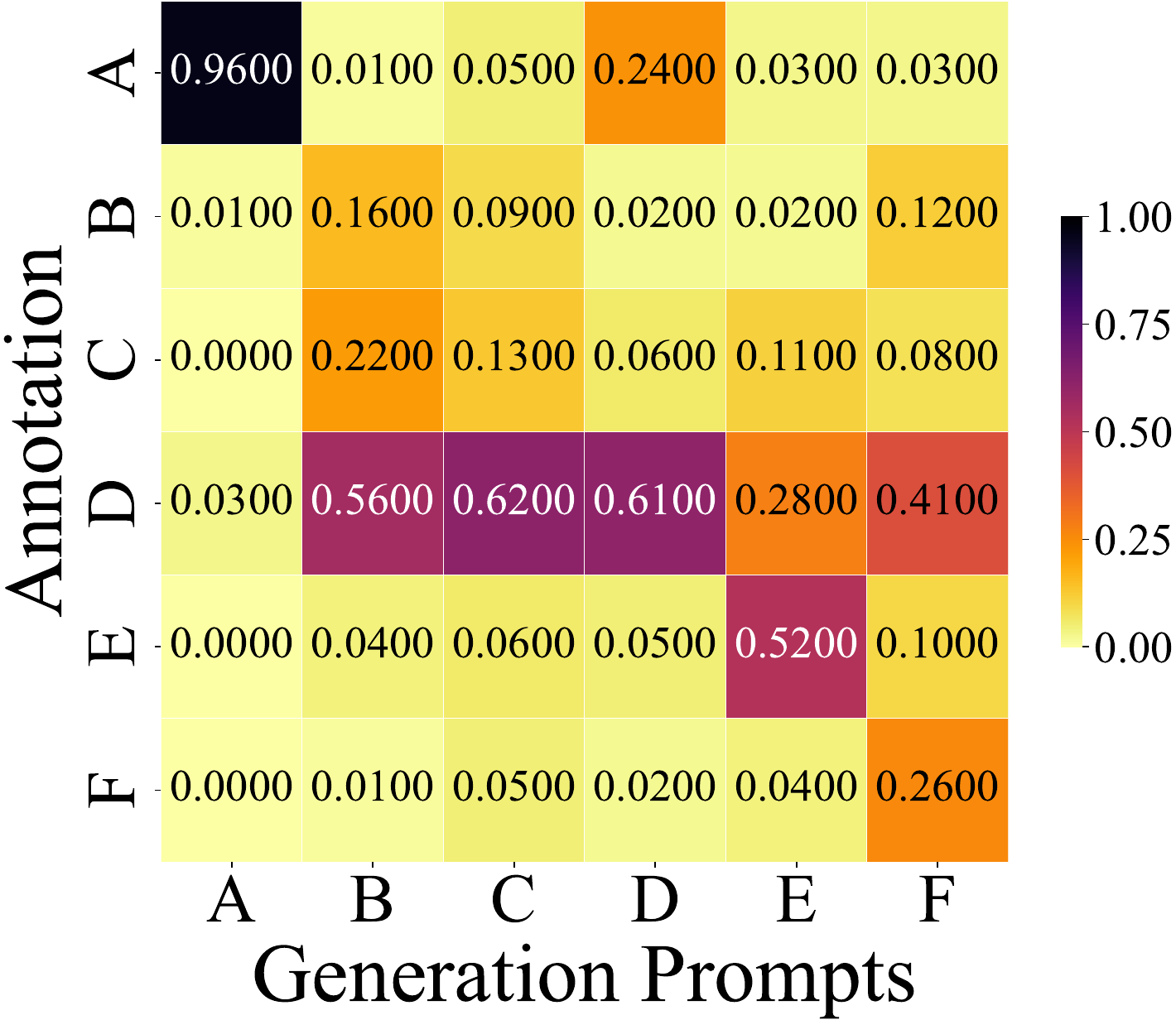}}
    \subfloat[ODE]{\includegraphics[width=0.245\linewidth]{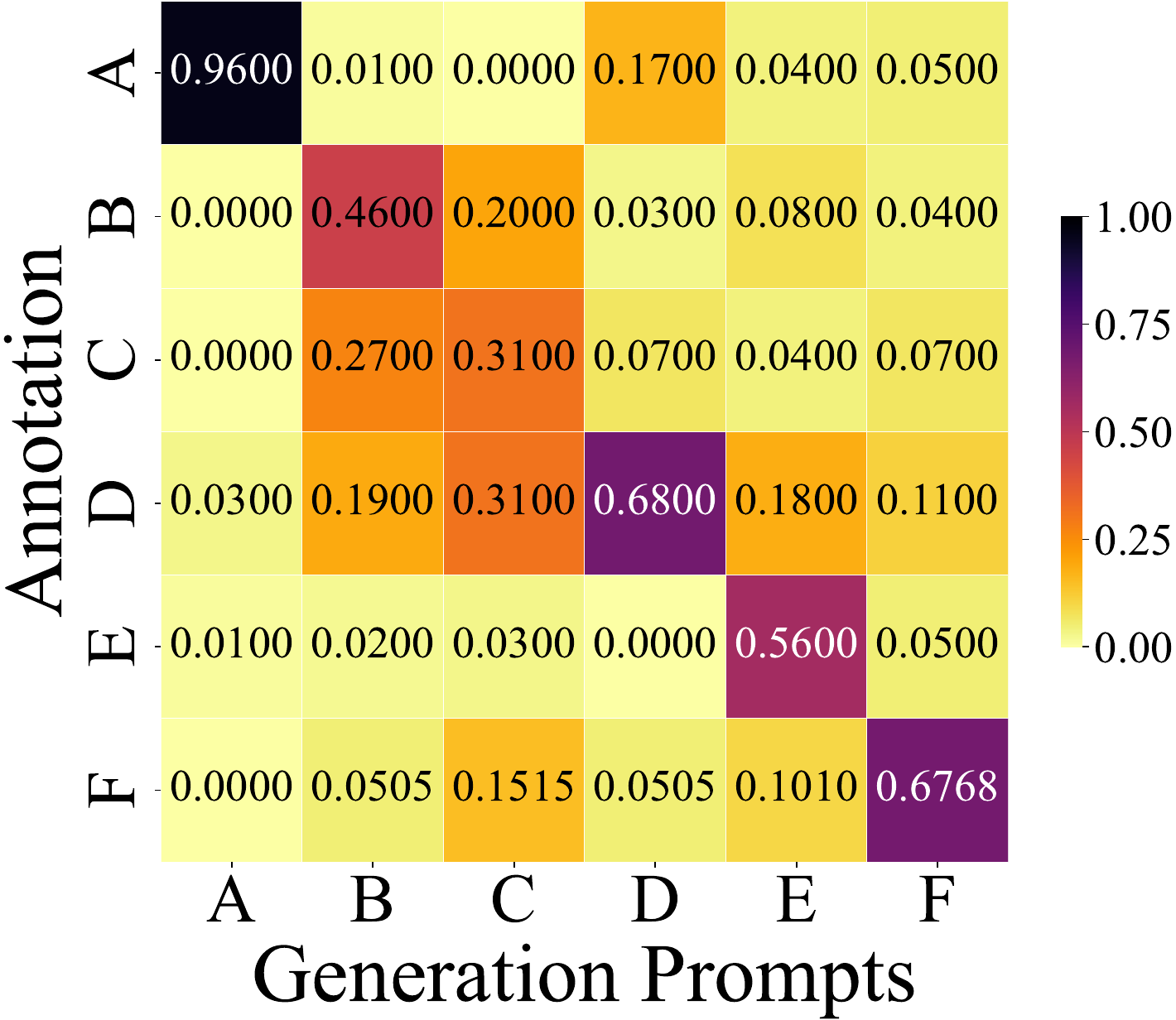}}
    \subfloat[PTS]{\includegraphics[width=0.245\linewidth]{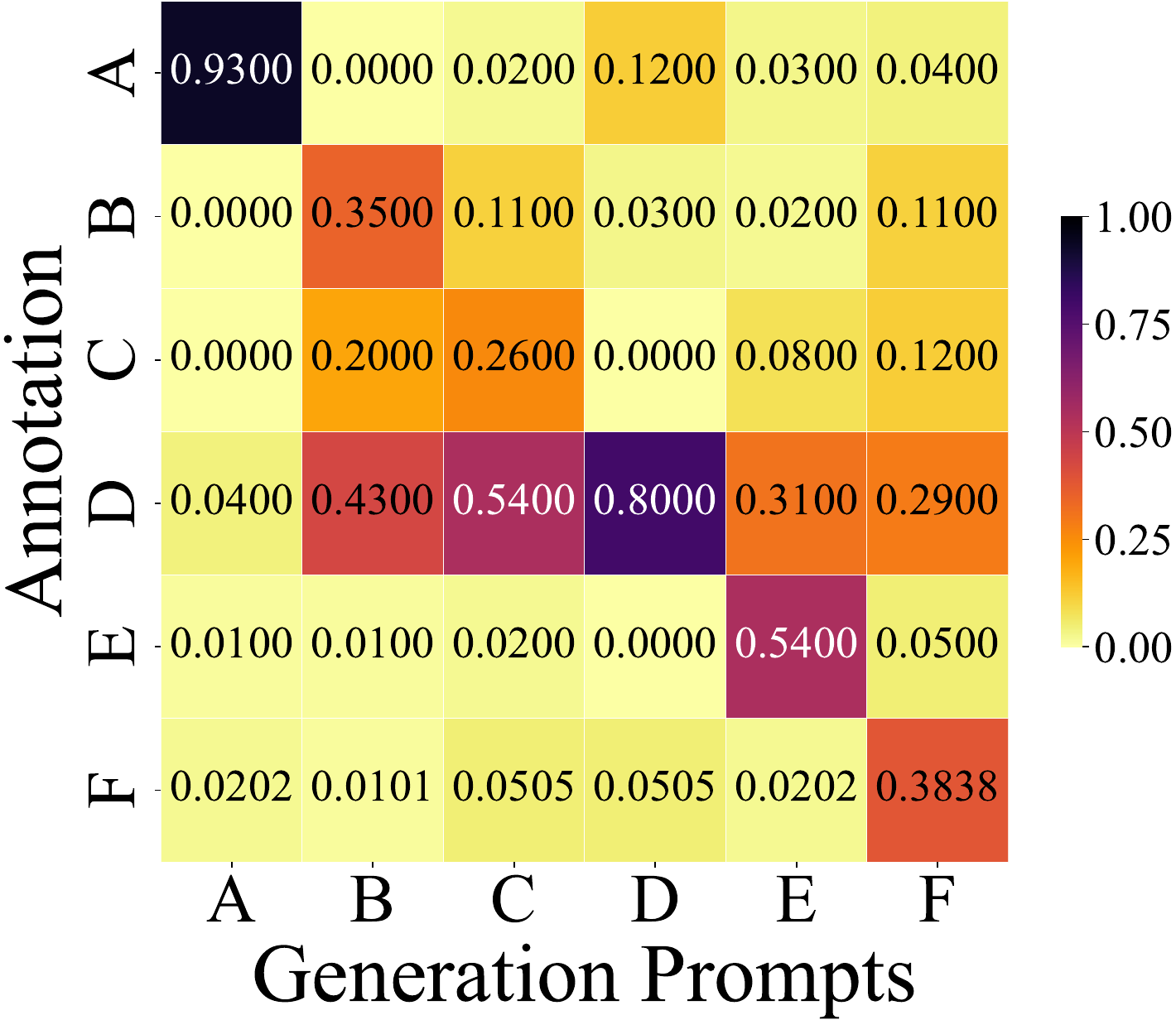}}
    \subfloat[SD1]{\includegraphics[width=0.245\linewidth]{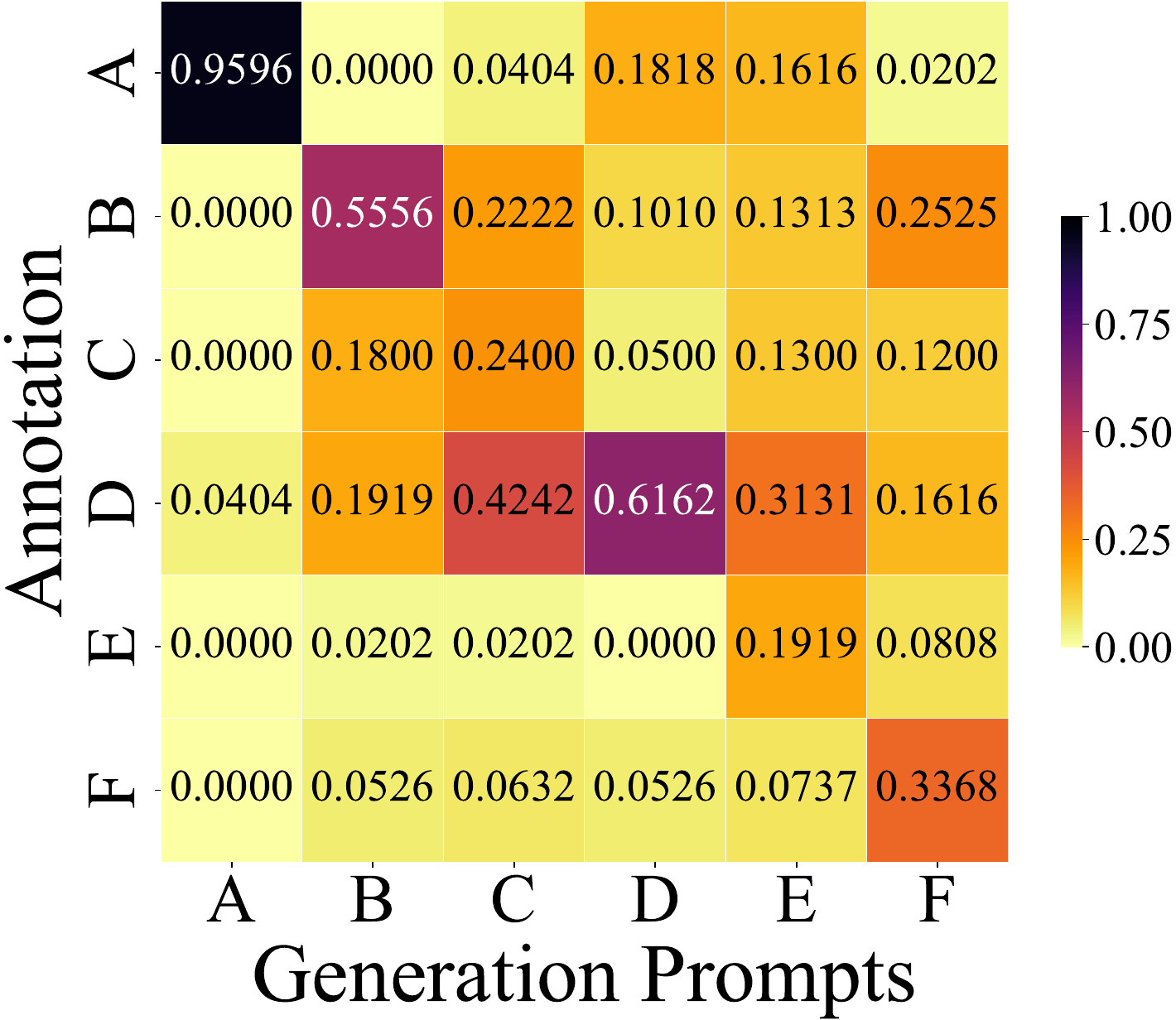}}

    \subfloat[SD2]{\includegraphics[width=0.245\linewidth]{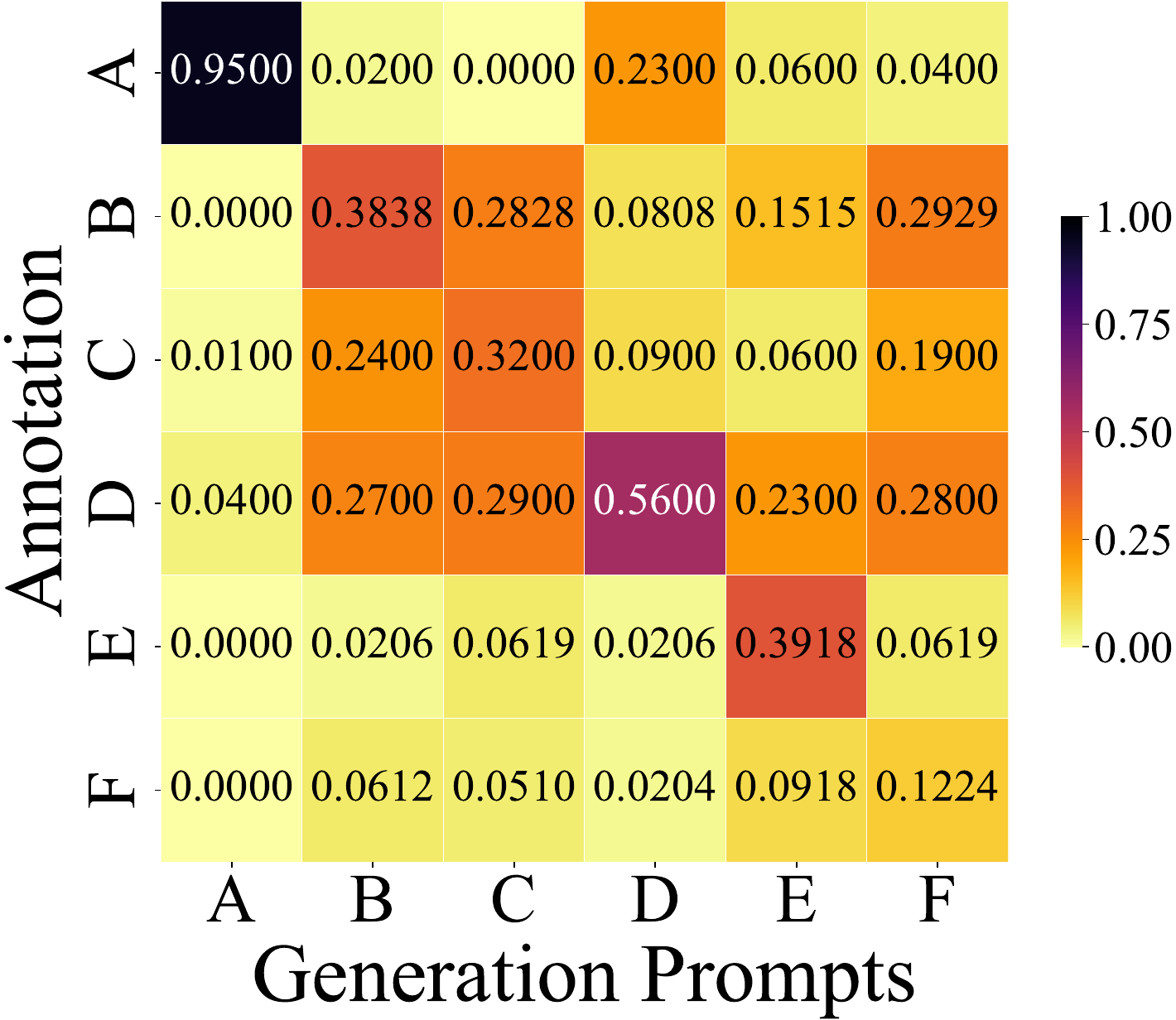}}
    \subfloat[SDX]{\includegraphics[width=0.245\linewidth]{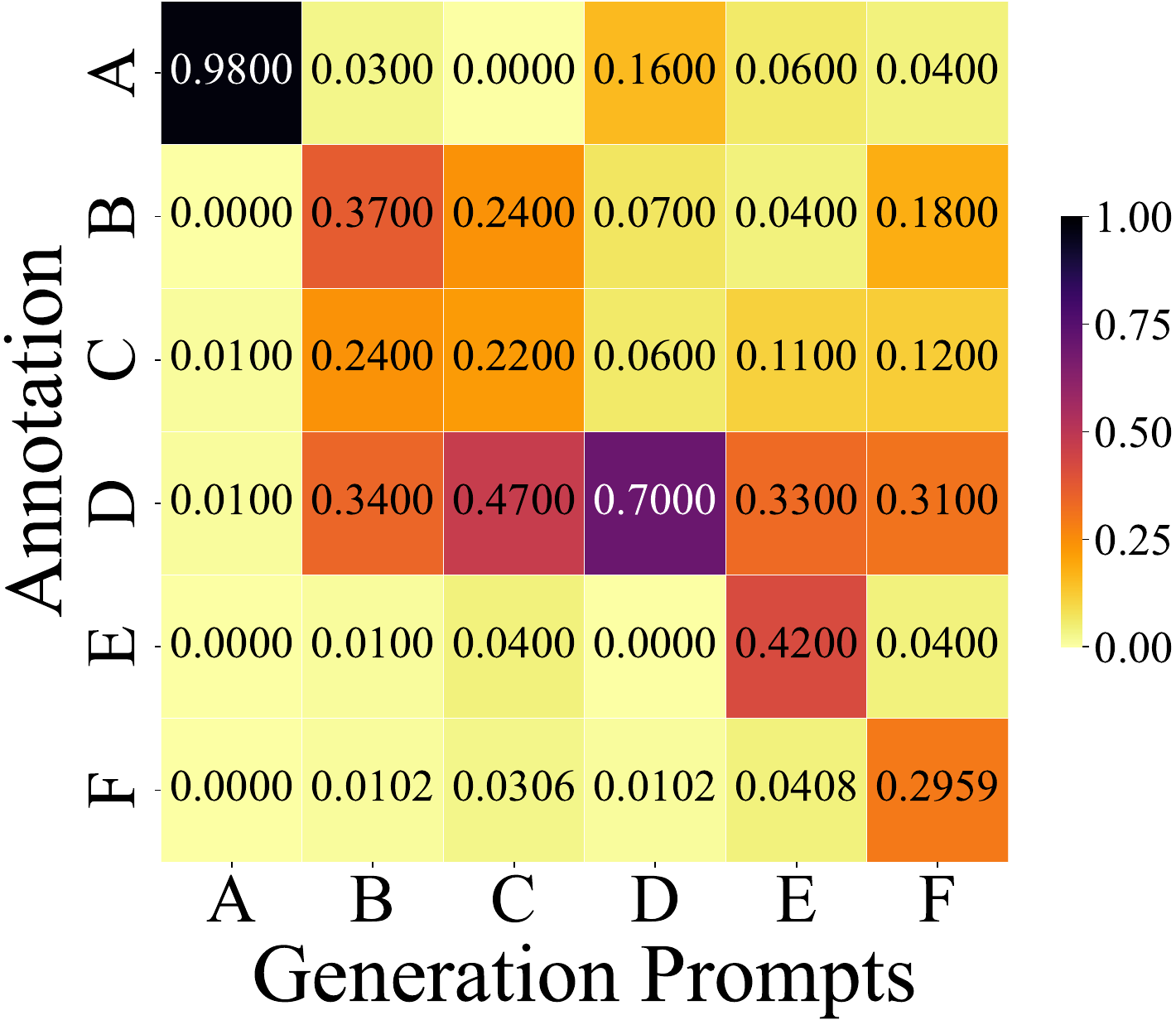}}
    \subfloat[SD3]{\includegraphics[width=0.245\linewidth]{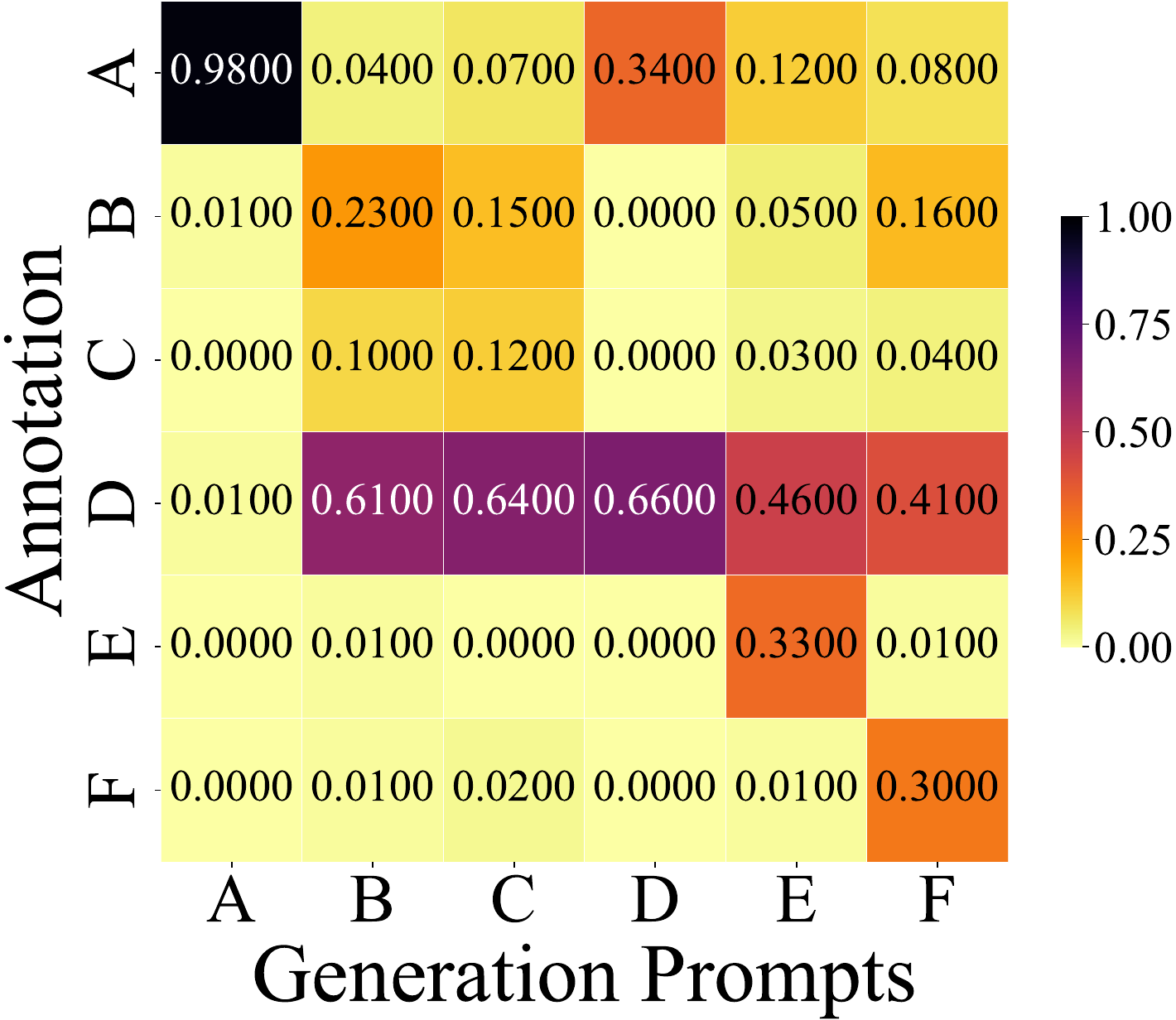}}
    \subfloat[SGA]{\includegraphics[width=0.245\linewidth]{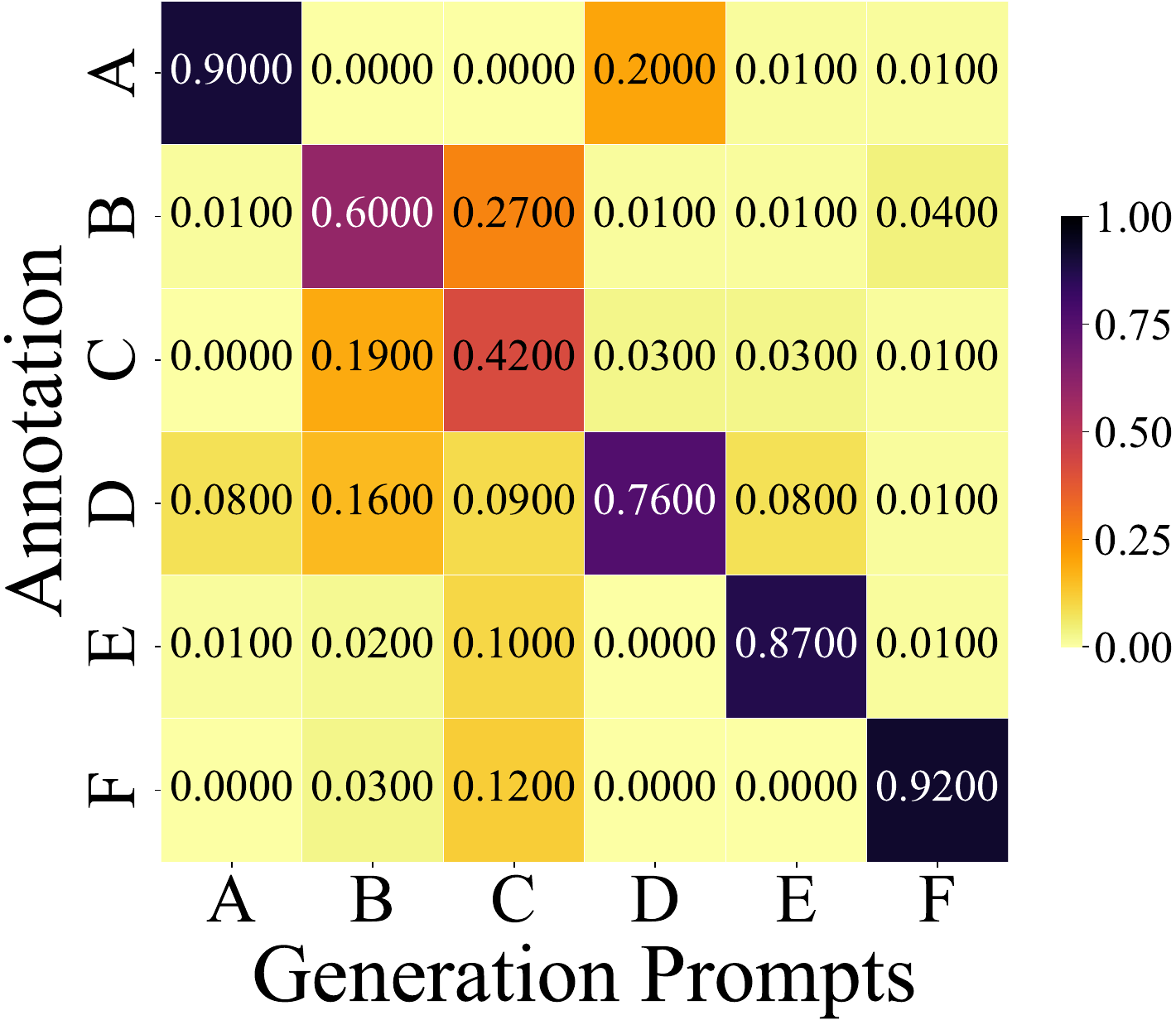}}
    
    
    \vspace{-0cm}
    \caption{Performance of different T2I models on the MEMO-Bench dataset for sentiment generation capability. Options A-F correspond to happiness (HAP), sadness (SAD), worry (WOR), neutrality (NEU), surprise (SUR) and anger (ANG), respectively.}
    \label{fig:res_t2i}
    \vspace{-0cm}
\end{figure*}

\subsection{Generated Image Quality}
\label{sec:quality}
To conduct a thorough evaluation of the quality of AGPIs within the MEMO-Bench, a detailed analysis of the distribution of Mean Opinion Scores (MOSs), derived from subjective assessments, is presented in Fig.~\ref{fig:mosdis}. The results in Fig.~\ref{fig:mosdis} illustrate a predominant trend where the majority of AGIs exhibit a high standard of visual fidelity. This finding not only highlights the outstanding performance of the various T2I models employed but also validates the ability of the AGPIs within MEMO-Bench to accurately and effectively convey character emotions, without being hindered by image quality issues. Further, to enable a more nuanced comparison of image quality across different T2I models and to examine how various emotional prompts influence image fidelity, the distribution of MOSs pertaining to these dual factors is elucidated in Fig.~\ref{fig:mosviolin}, which reveals several key observations: 1) There are substantial differences in image quality across the outputs of different T2I models, with the FCS model producing the highest quality images, while earlier T2I models, such as SD1 and SD2, exhibit relatively lower quality; 2) The emotional prompts used have varying impacts on the quality of the generated images. Specifically, images generated from neutral emotion prompts tend to have the highest quality, suggesting that existing T2I models are more adept at generating general facial representations. This observation points to a limitation in current T2I methods with regard to emotion generation; 3) A noticeable difference in image quality between positive and negative emotional prompts is evident, with images generated from positive emotion prompts being of superior quality compared to those generated from negative emotions. This discrepancy indicates that generating high-quality AGPIs in response to negative emotional prompts remains a significant challenge for current T2I models.

\subsection{Emotion Generation Capacity}
\label{sec:gen}
Two metrics, $GACC$ and $GERR$, are computed for all AGIs in the MEMO-Bench dataset, and the results are presented in Table~\ref{tab:t2iper} and Fig.~\ref{fig:res_t2i}, from which some insights can be drawn: 1) There are significant variations in the emotion generation capabilities of different T2I models. Specifically, SGA outperforms all other models in terms of sentiment generation accuracy, whereas FCS exhibits the poorest performance, with a gap of over 35\% in GACC between these two models; 2) In terms of sentiment categories, T2I models generally demonstrate strong sentiment generation ability for positive emotions, such as HAP and NEU, but show limited capability in generating negative emotions, which is consistent with the findings in Sec.~\ref{sec:quality}; 3) As shown in Fig.~\ref{fig:res_t2i}, regions B, C, and D are predominantly darker in color, indicating that the existing T2I models tend to confuse the generation of emotions such as SAD, WOR, and NEU. Overall, the sentiment generation performance of most T2I models remains suboptimal.


%% file: sec/5_bench2.tex
\begin{figure*}[!t]
    \centering
    \subfloat[GPT-4o]{\includegraphics[width=0.245\linewidth]{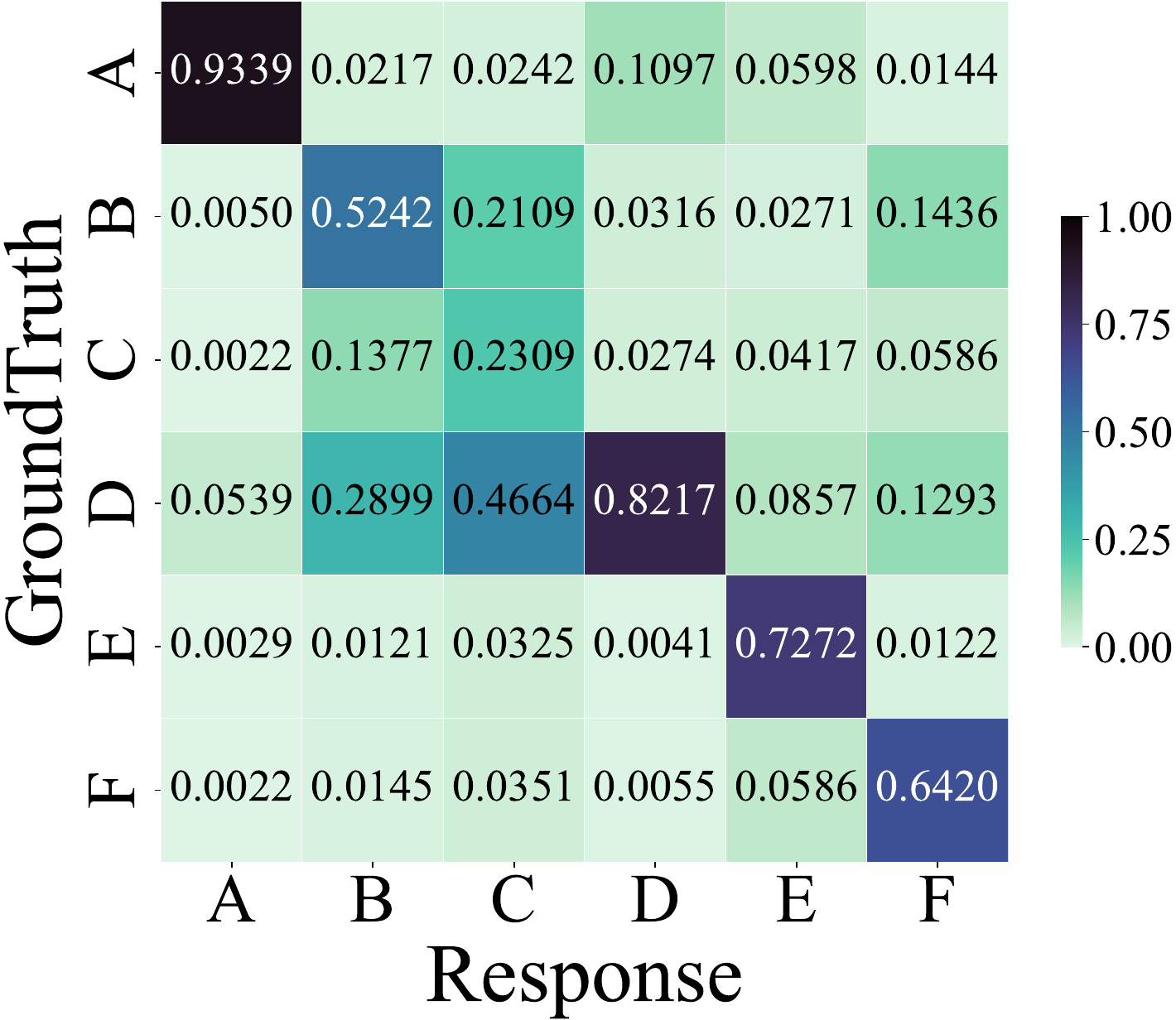}}
    \subfloat[Gemini-1.5-Pro]{\includegraphics[width=0.245\linewidth]{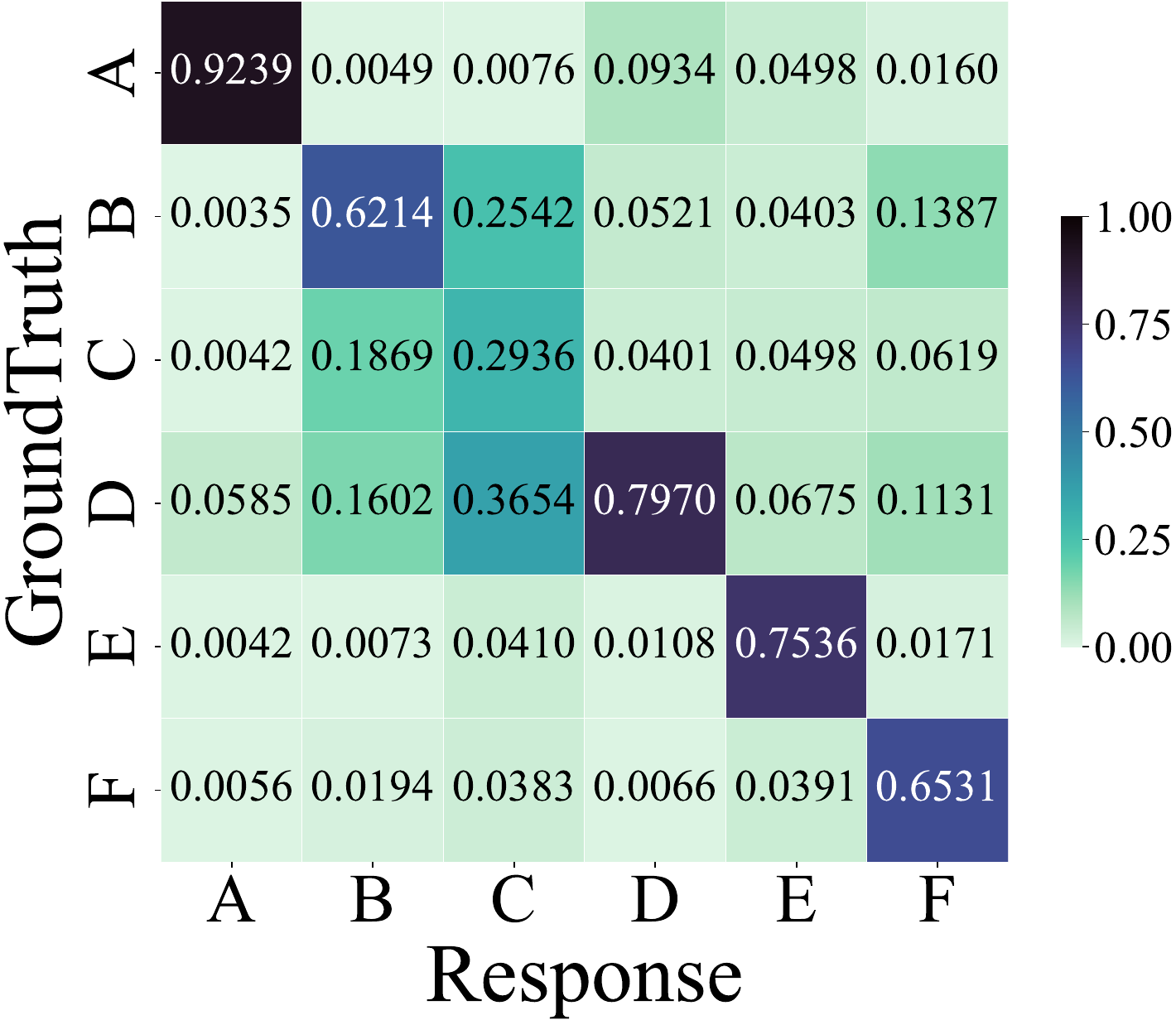}}
    \subfloat[IDEFICS-Instruct]{\includegraphics[width=0.245\linewidth]{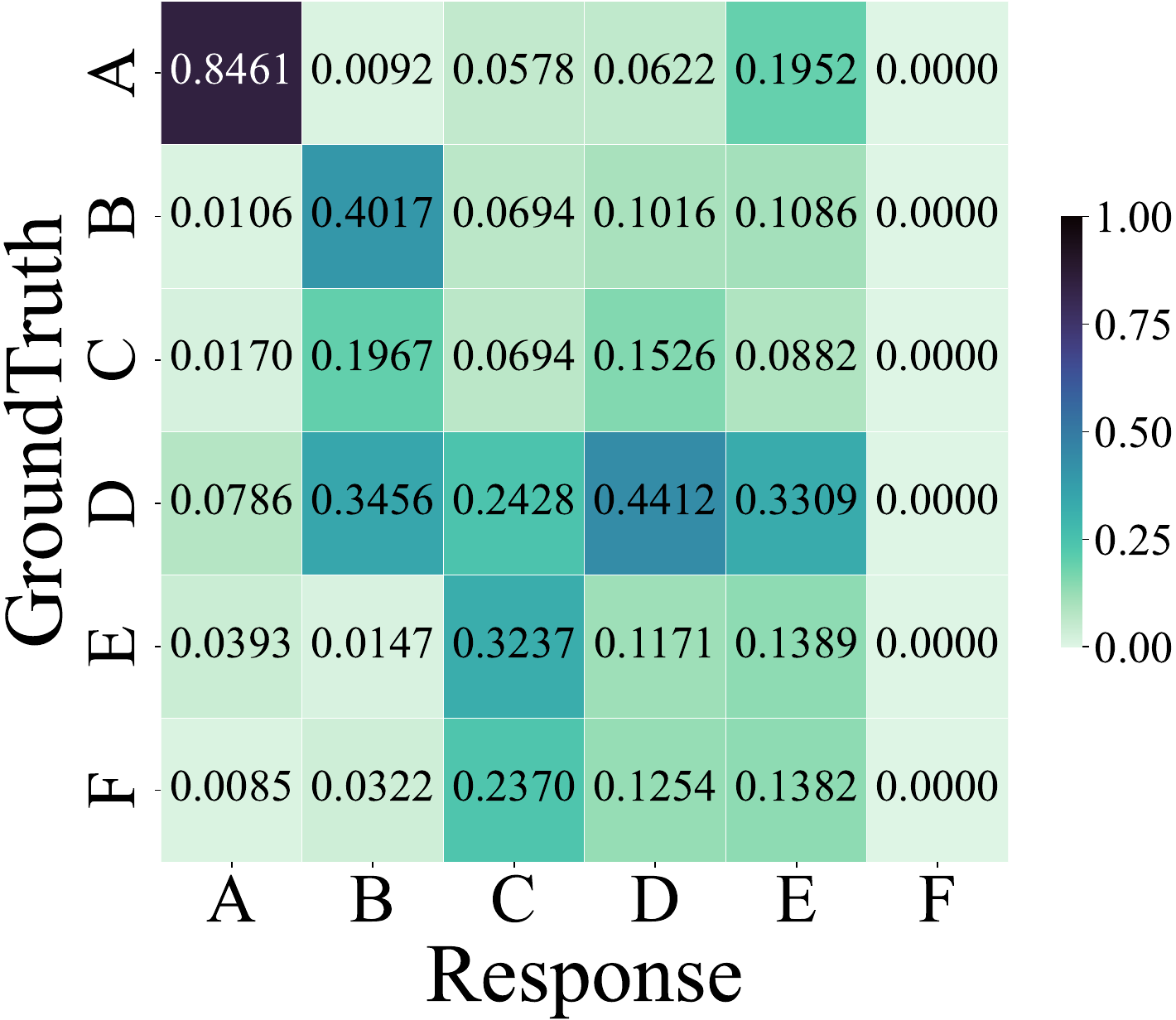}}
    \subfloat[InstructBLIP (Flan-T5-XL)]{\includegraphics[width=0.245\linewidth]{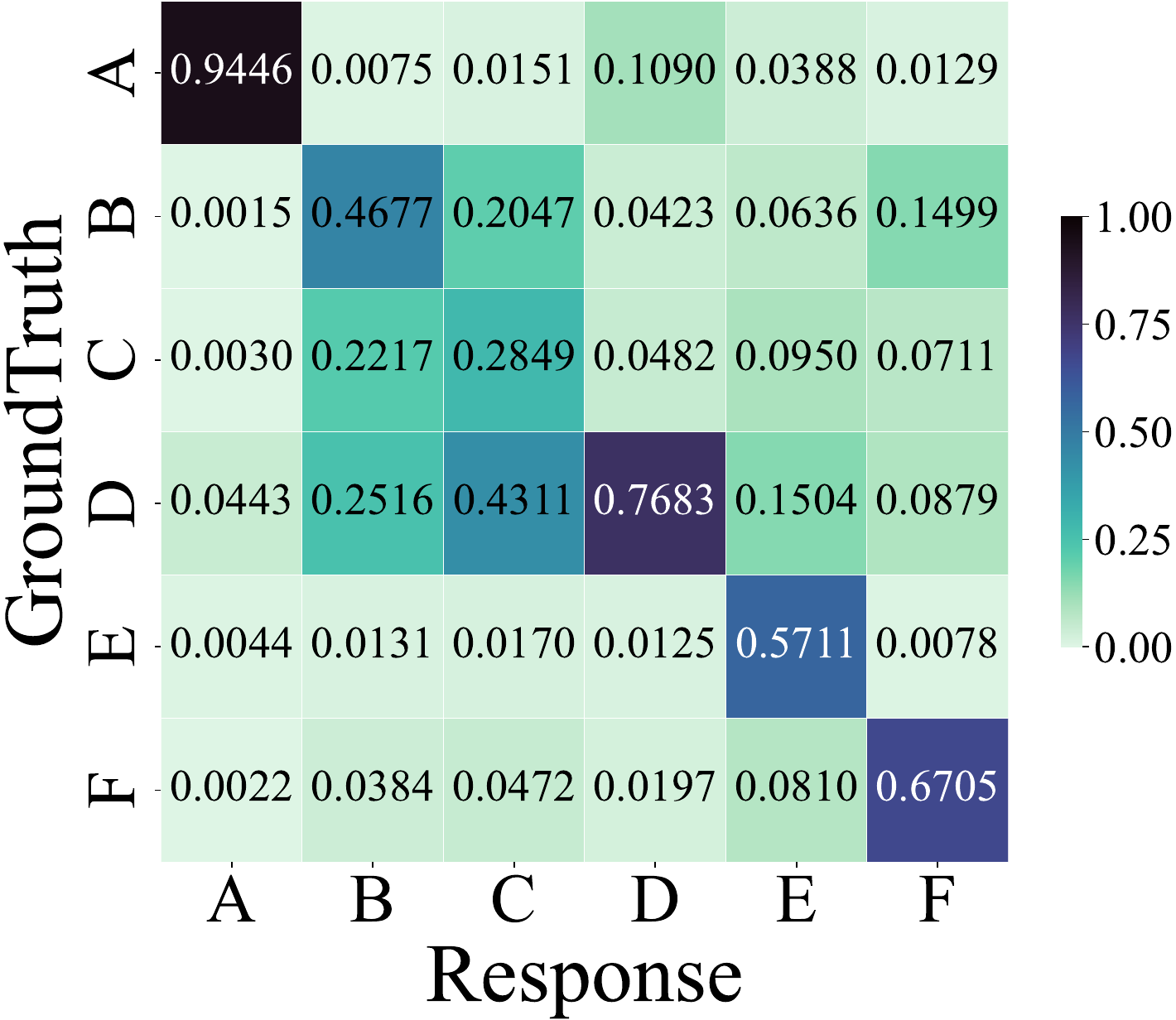}}
    
    \subfloat[InstructBLIP (Vicuna-7B)]{\includegraphics[width=0.245\linewidth]{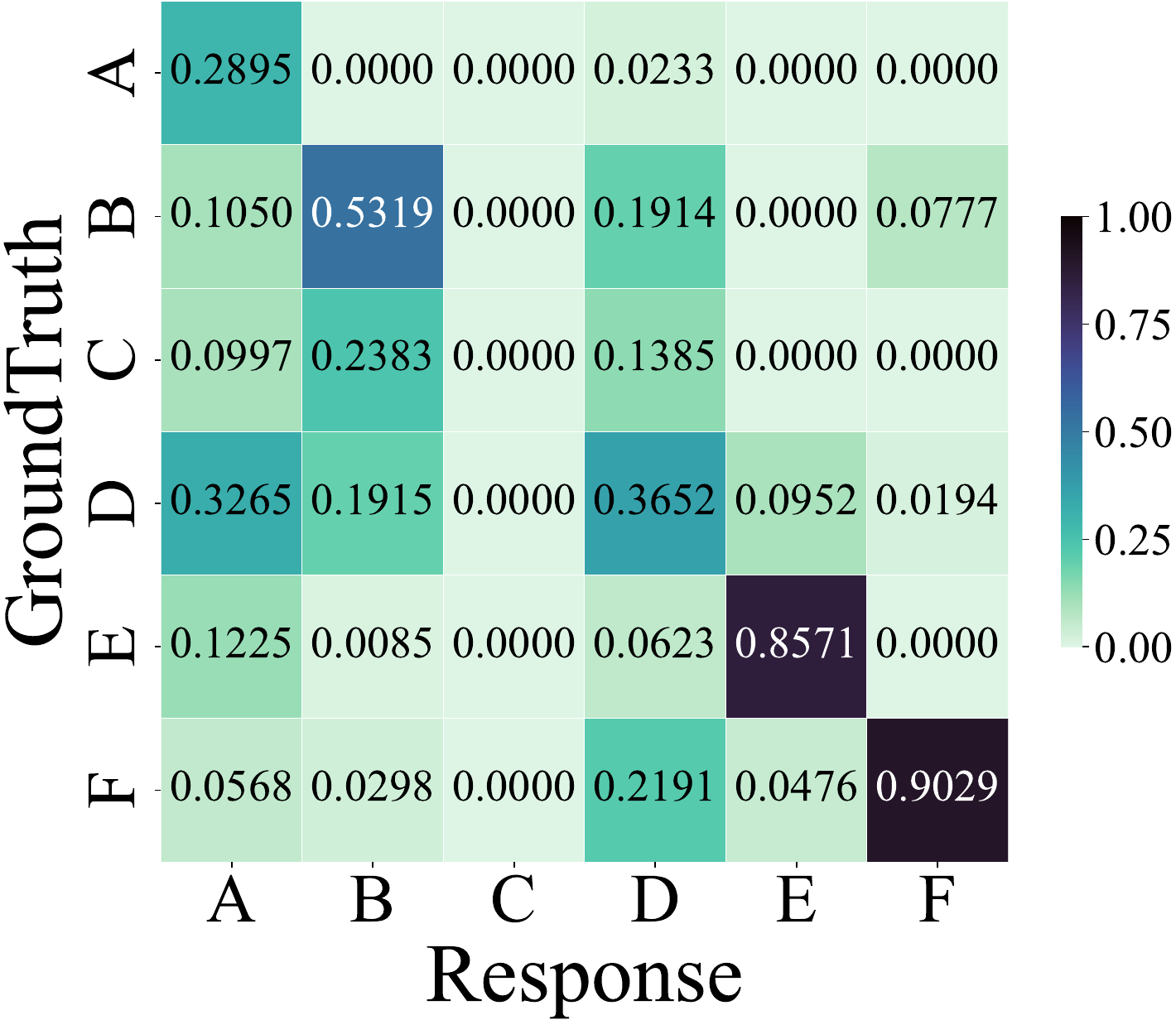}}
    \subfloat[InternLM-XComposer-VL]{\includegraphics[width=0.245\linewidth]{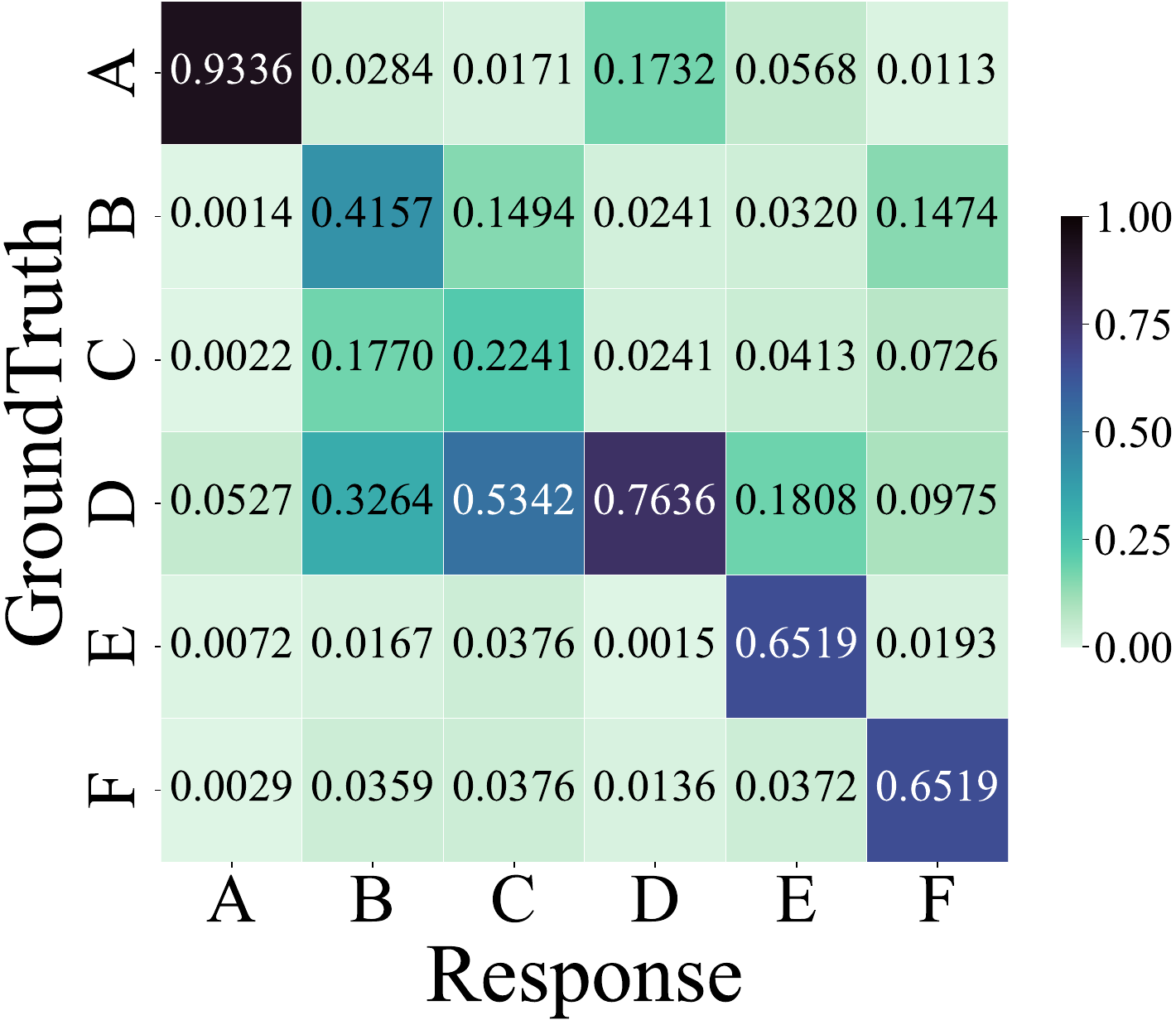}}
    \subfloat[LLaVa-v1.5-7B]{\includegraphics[width=0.245\linewidth]{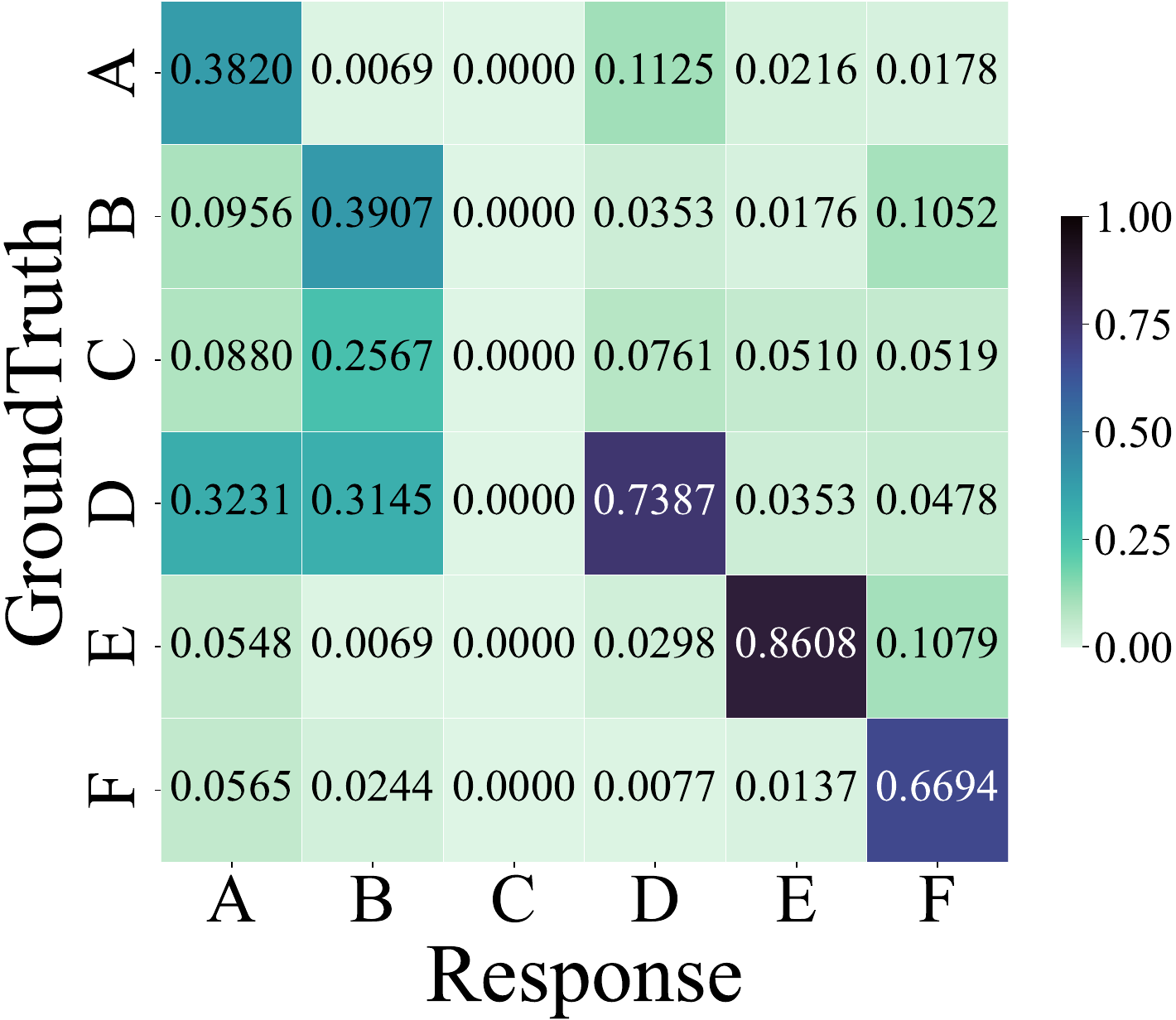}}
    \subfloat[LLaVa-v1.5-13B]{\includegraphics[width=0.245\linewidth]{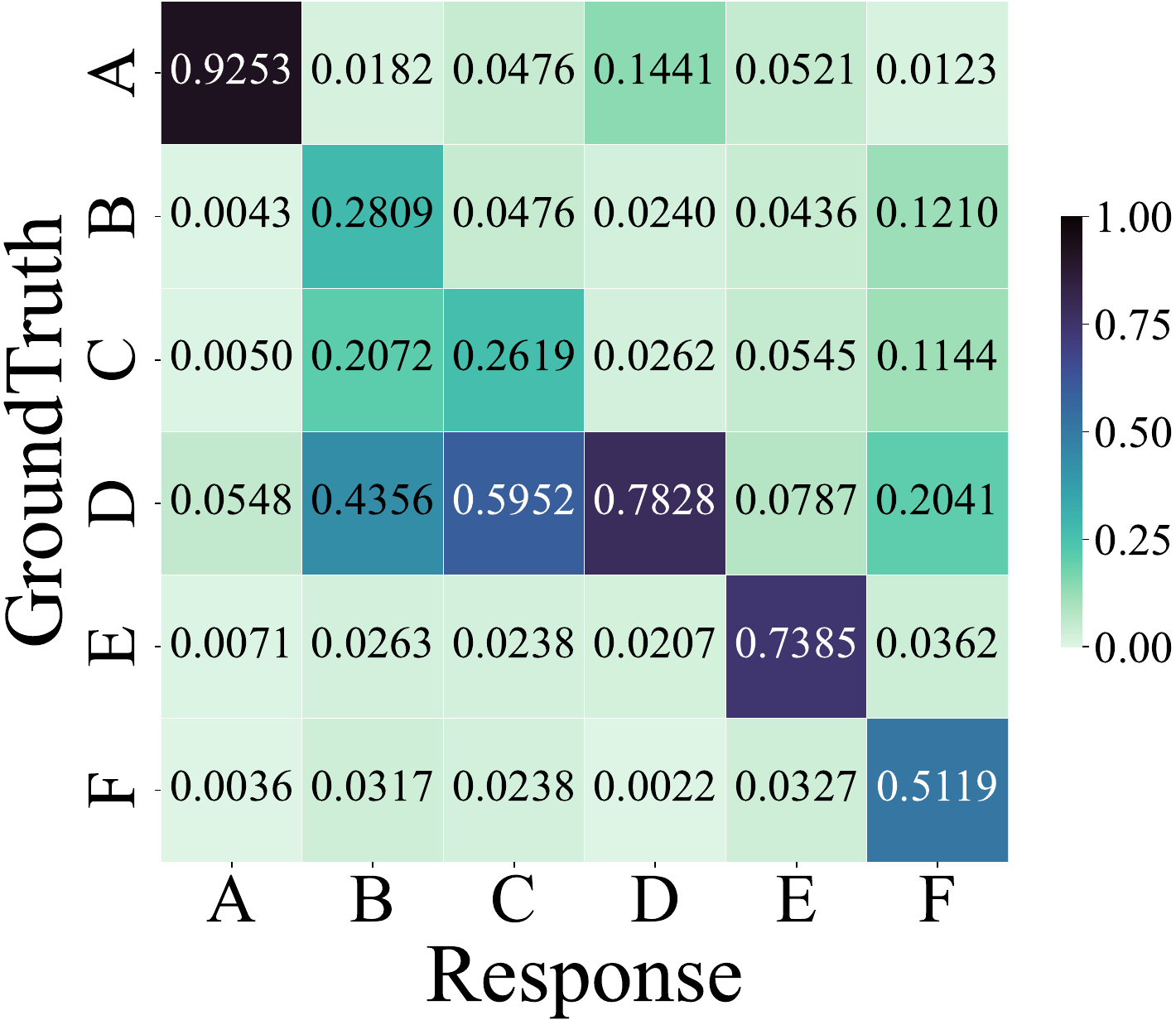}}

    \subfloat[LLaVa-v1.6-13B]{\includegraphics[width=0.245\linewidth]{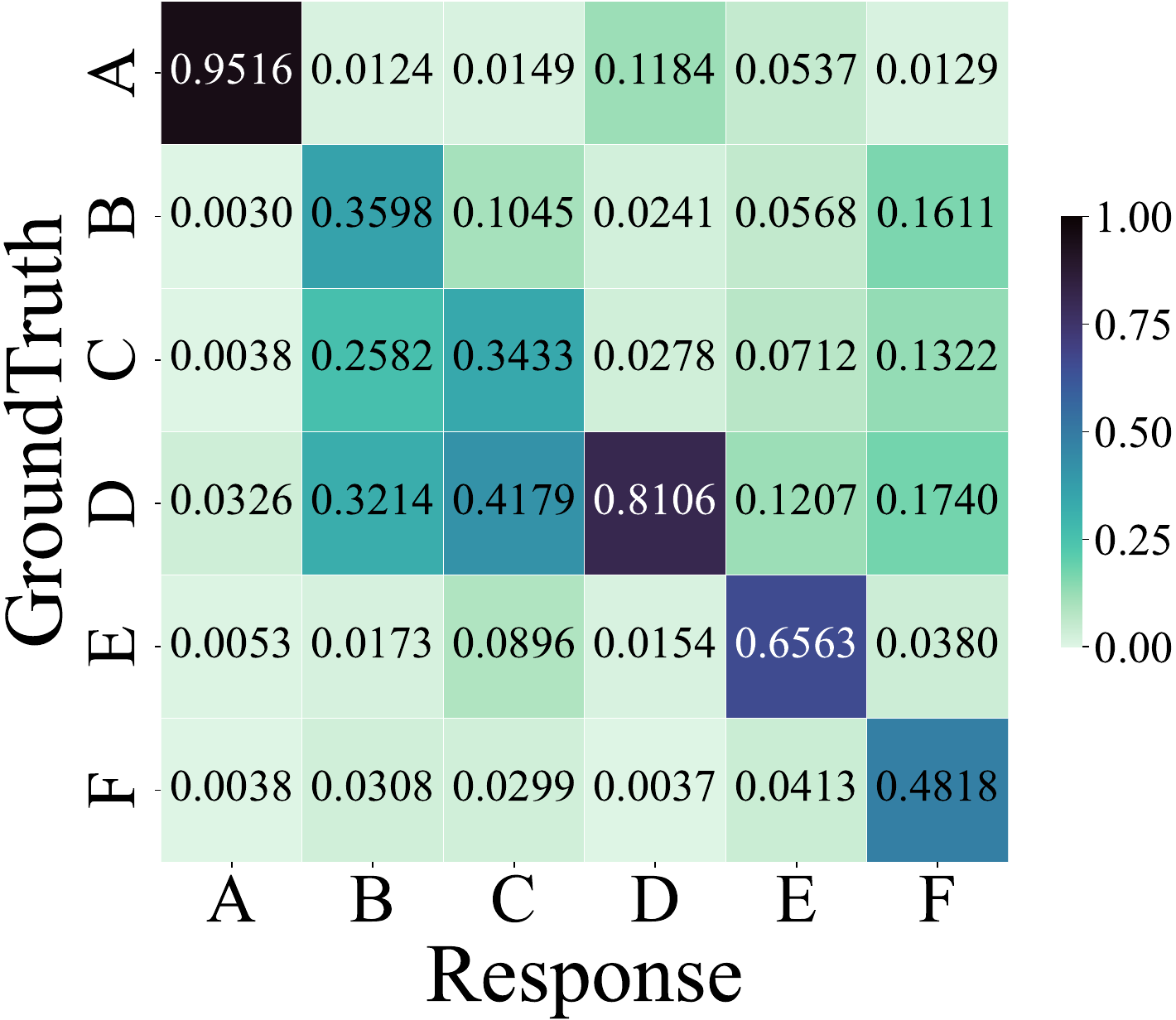}}
    \subfloat[mPLUG-Owl1]{\includegraphics[width=0.245\linewidth]{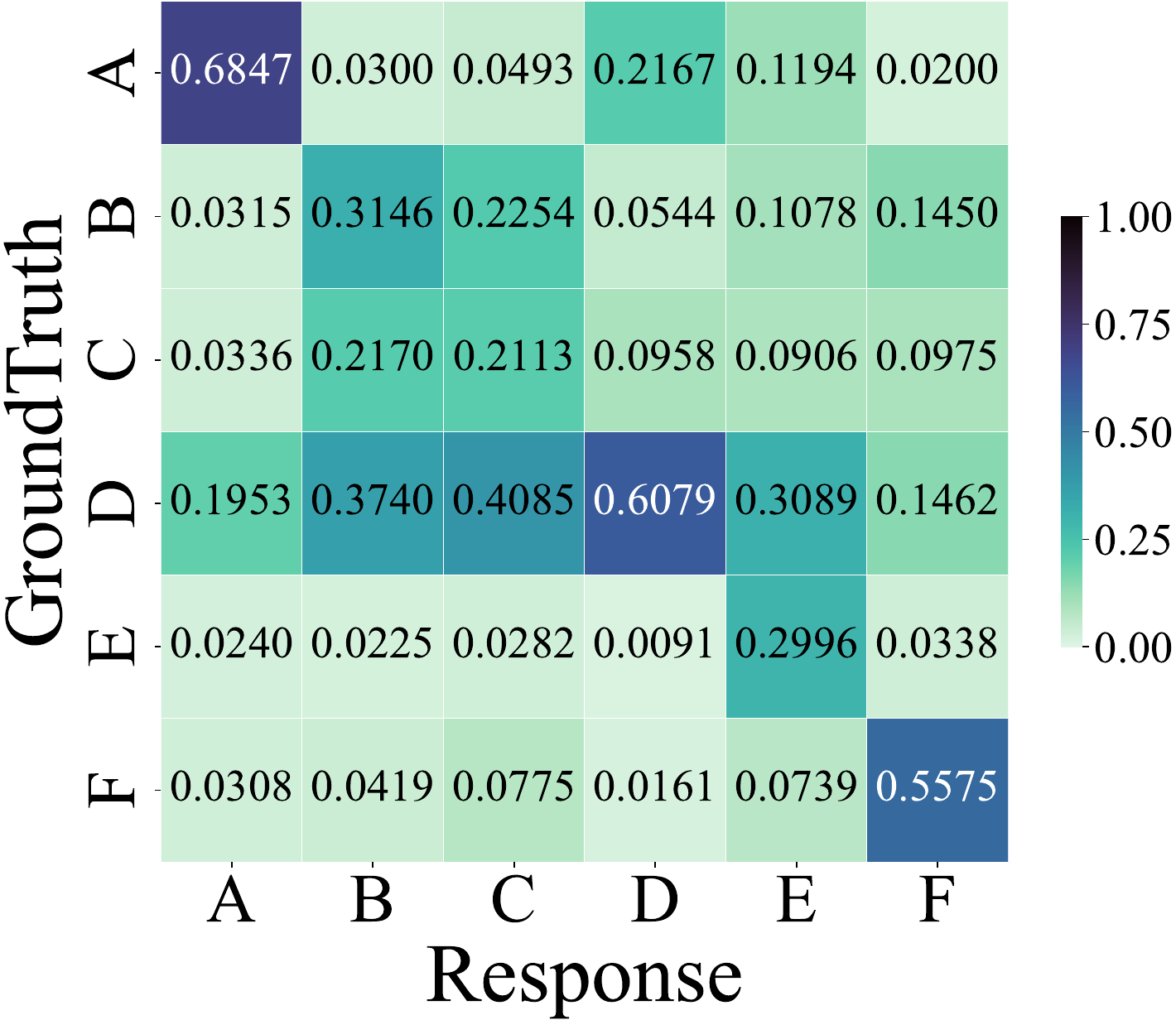}}
    \subfloat[mPLUG-Owl2]{\includegraphics[width=0.245\linewidth]{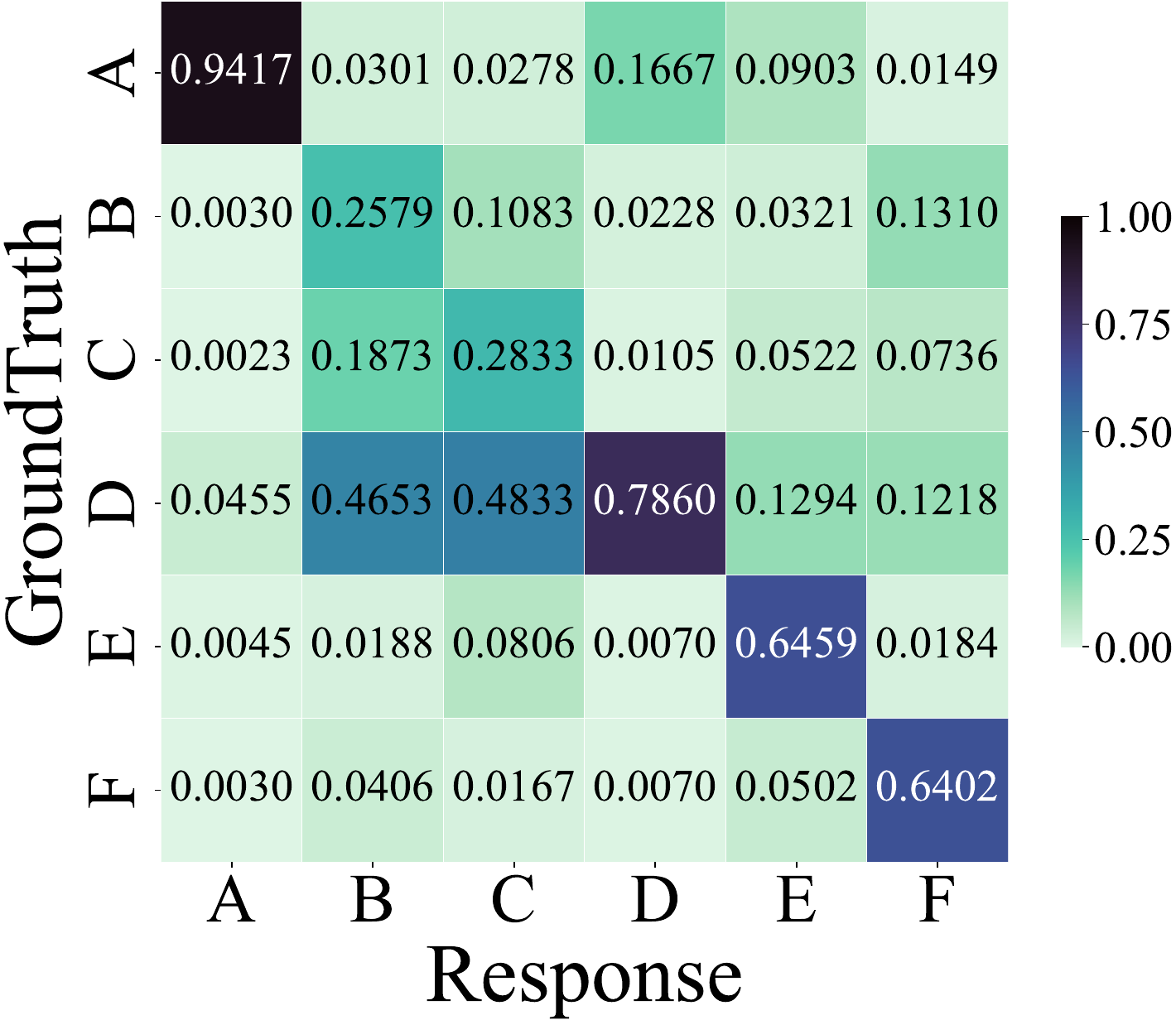}}
    \subfloat[mPLUG-Owl3]{\includegraphics[width=0.245\linewidth]{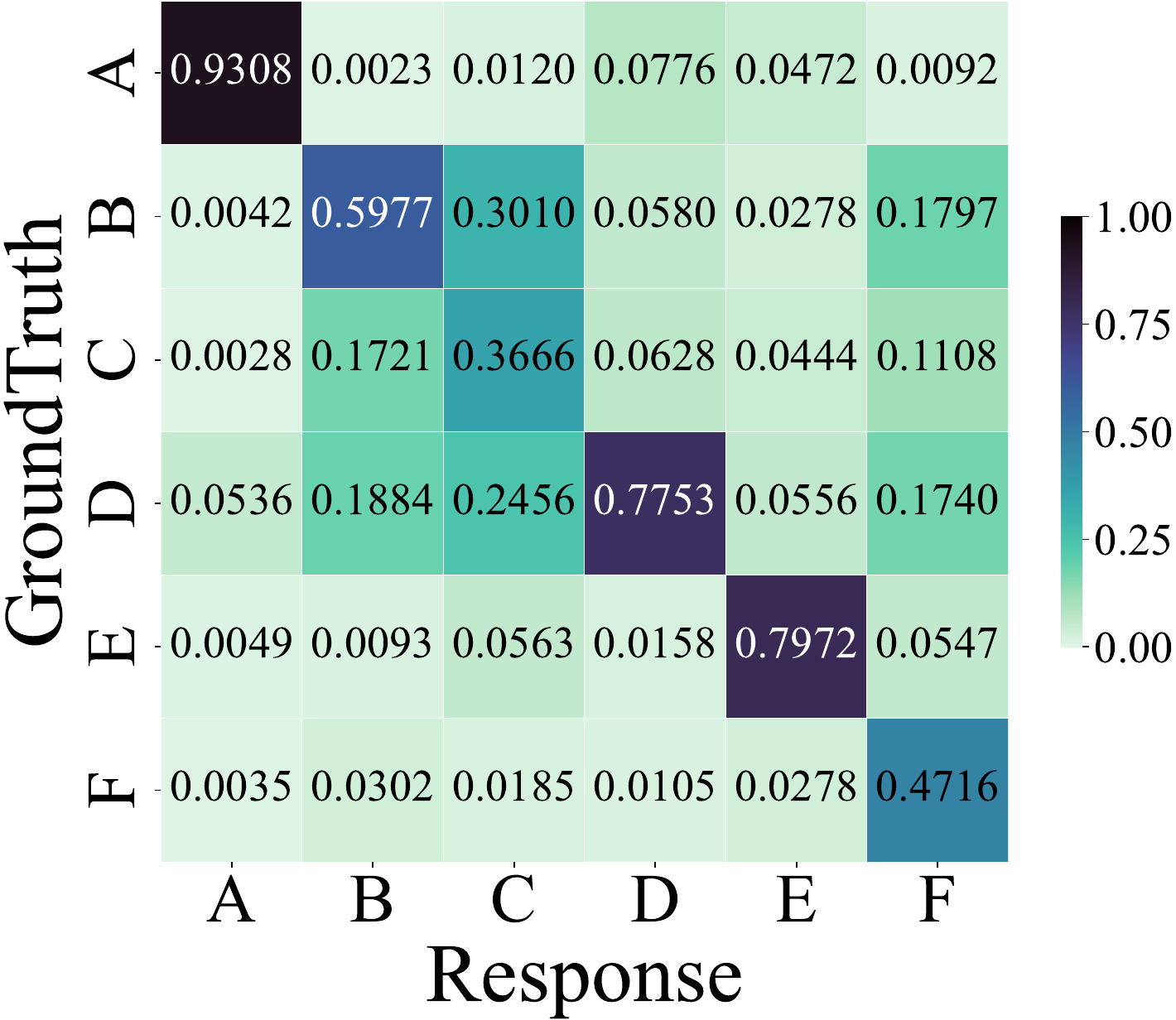}}

    \subfloat[Otter-v1 (MPT-7B)]{\includegraphics[width=0.245\linewidth]{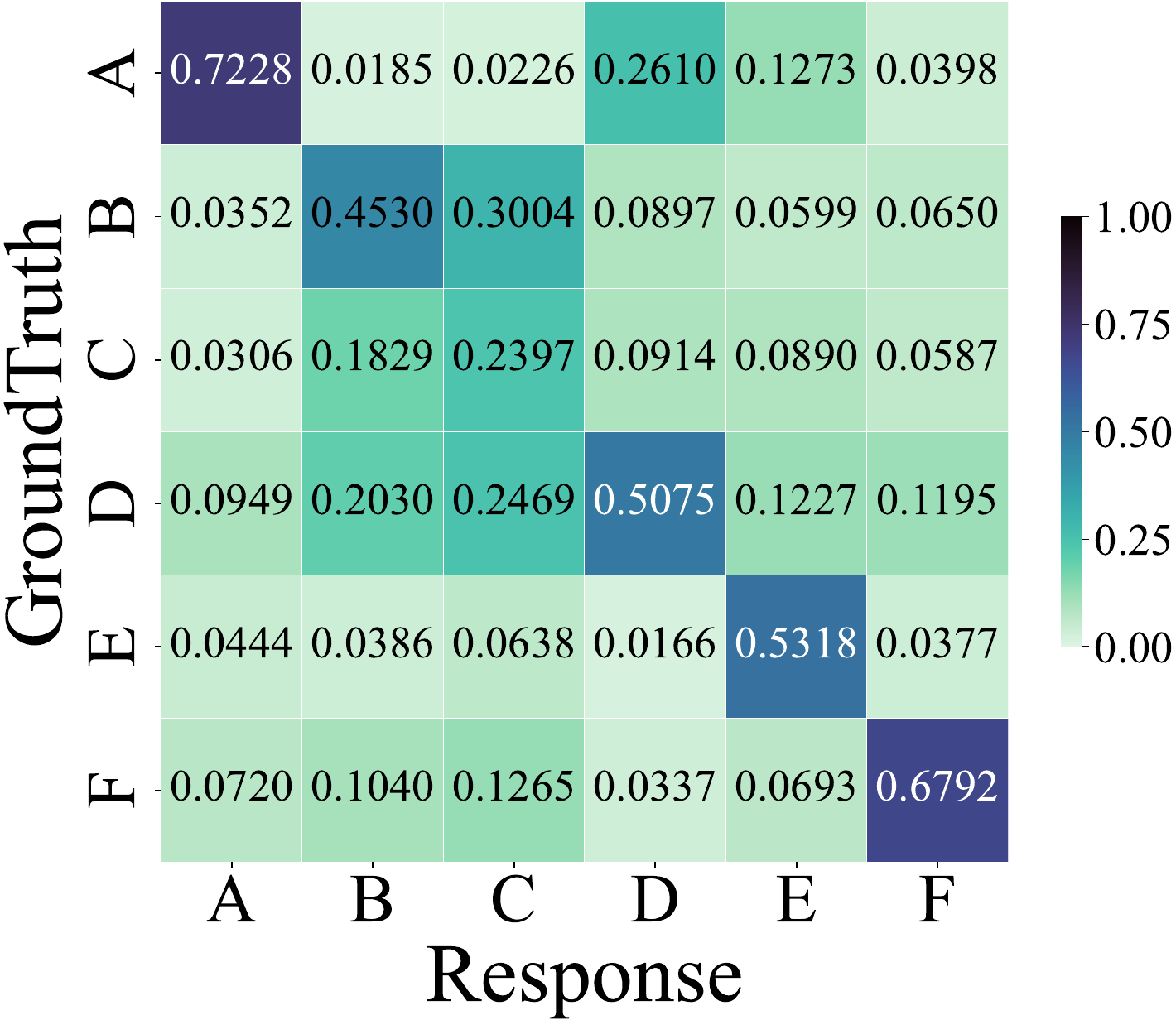}}
    \subfloat[Qwen-VL]{\includegraphics[width=0.245\linewidth]{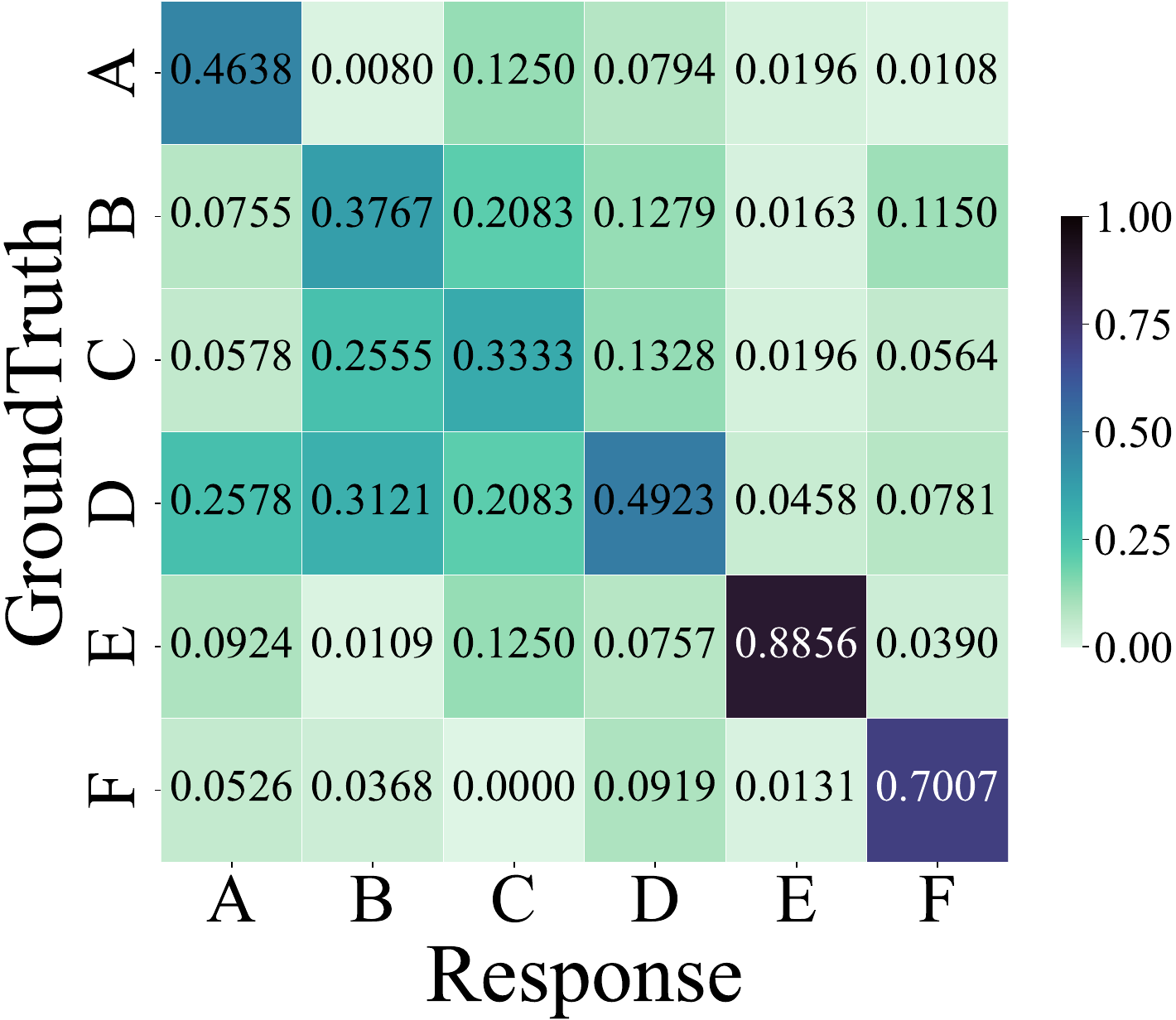}}
    \subfloat[Shikra (Vicuna-7B)]{\includegraphics[width=0.245\linewidth]{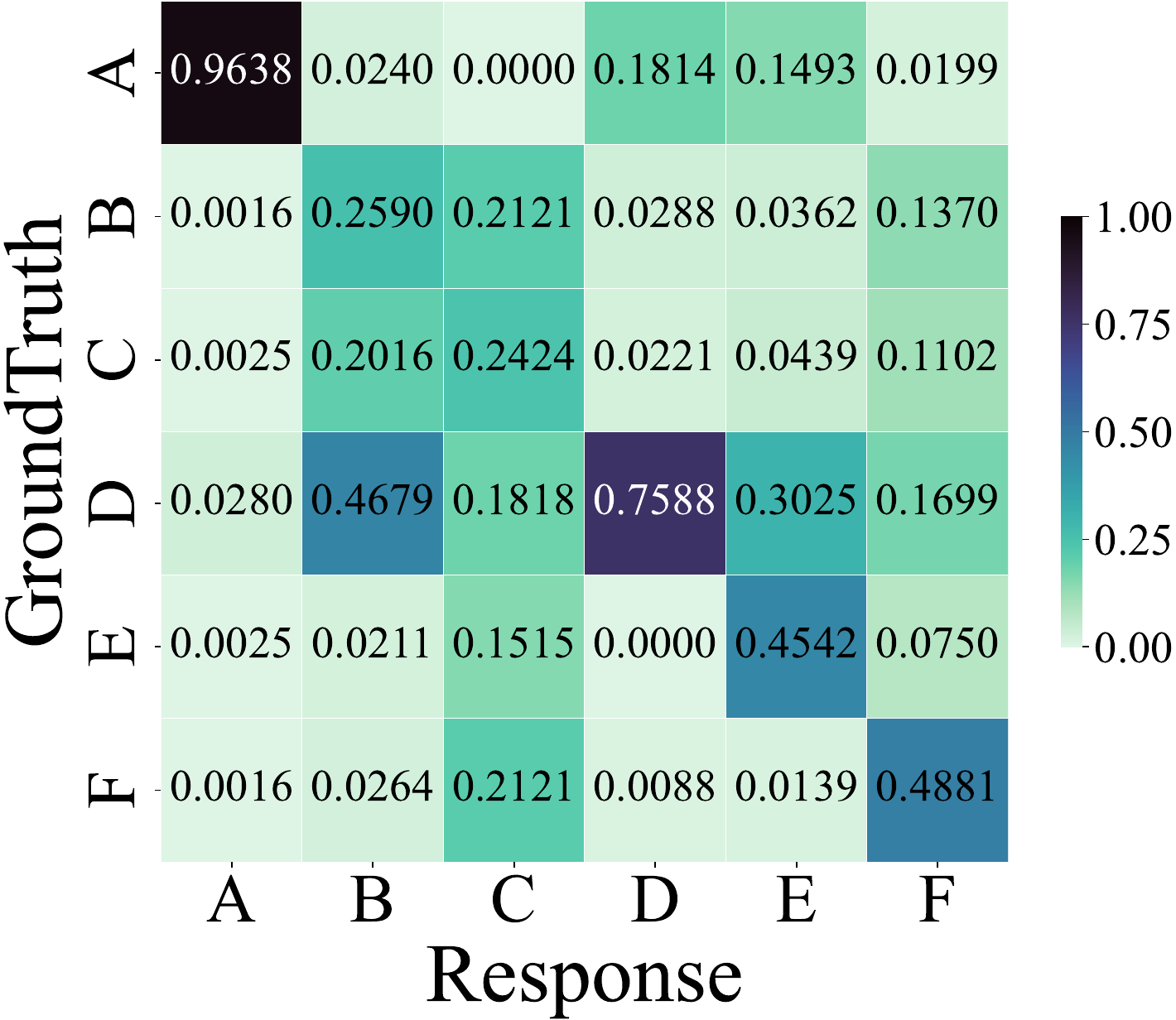}}
    \subfloat[VisualGLM-6B]{\includegraphics[width=0.245\linewidth]{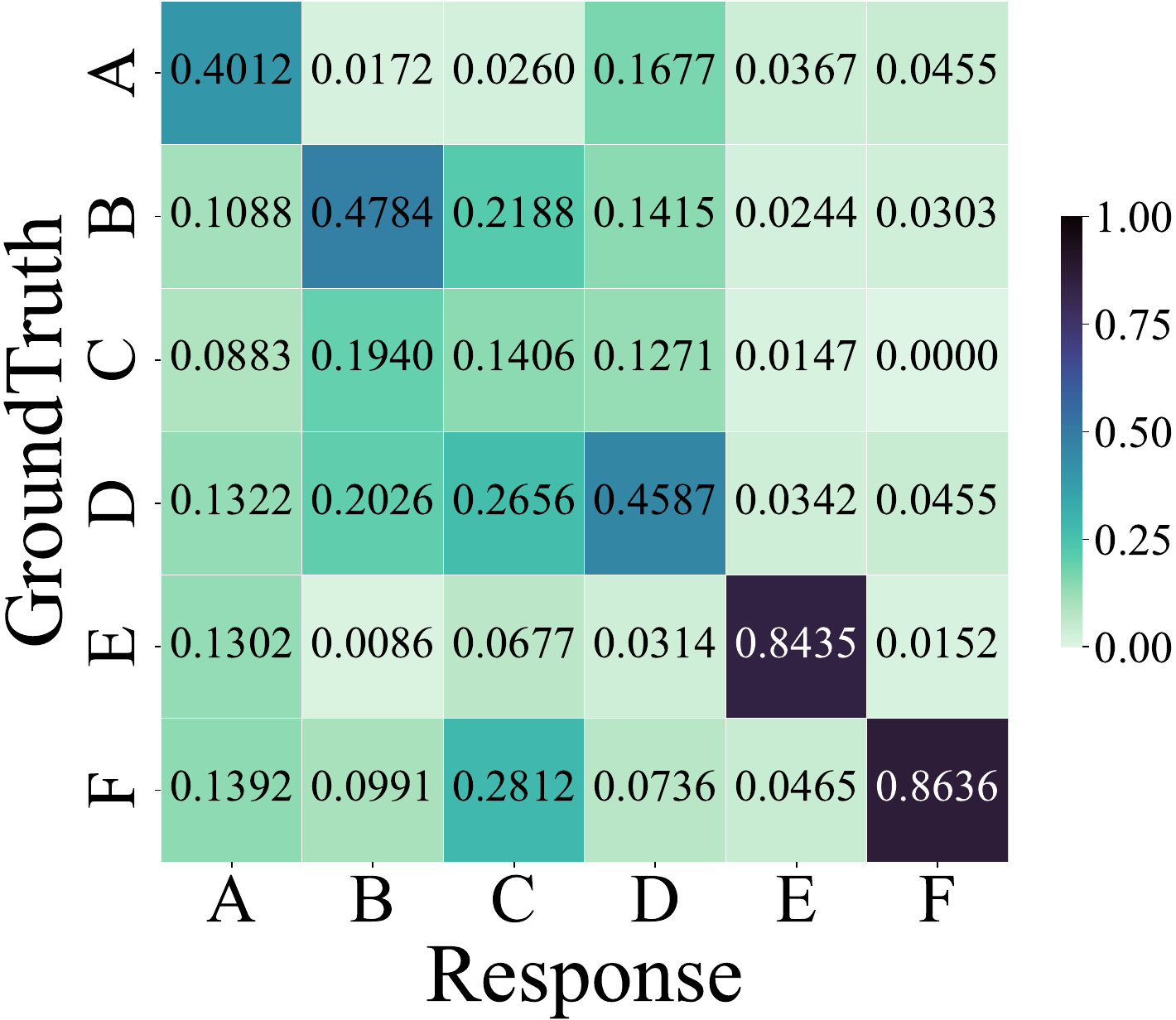}}
    
    \vspace{-0cm}
    \caption{Performance of different MLLMs on the MEMO-Bench dataset for emotional comprehension. Options A-F correspond to happiness (HAP), sadness (SAD), worry (WOR), neutrality (NEU), surprise (SUR) and anger (ANG), respectively.}
    \label{fig:res_mllm}
    \vspace{-0.2cm}
\end{figure*}

\begin{table*}[th]
\centering
\renewcommand\tabcolsep{1.4pt} 
\caption{Benchmark for MLLMs in human emotion understanding. Best in {\bf\textcolor{red}{RED}}, second in \bf\textcolor{blue}{BLUE}.}
\vspace{-0.2cm}

\begin{tabular}{l|ccccccc|cccc}
\toprule
\multirow{2}{*}{MLLM}  & \multicolumn{7}{c|}{Coarse-grained stage ($UACC$)} & \multicolumn{4}{c}{Fine-grained stage} \\ 
\cline{2-12}
 & HAP& SAD& WOR       & NEU      & SUR     & ANG     & $All$& SRCC$\uparrow$      & PLCC$\uparrow$      & KRCC$\uparrow$       & RMSE$\downarrow$   \\ \hline
\rowcolor{gray!25}GPT-4o \cite{gpt4o}&0.9339 & 0.5242& 0.2309& \bf\textcolor{red}{0.8217}&0.7272 &0.6420 & 0.5669 &\bf\textcolor{red}{0.1802} &\bf\textcolor{blue}{0.2625} &\bf\textcolor{red}{0.1365} & 0.8092           \\
Gemini-1.5-Pro \cite{team2024gemini}& 0.9239& \bf\textcolor{red}{0.6214}& 0.2936& 0.7970& 0.7536& 0.6531& \bf\textcolor{blue}{0.6574}& 0.0664& 0.1439& 0.0501& 0.8188            \\
\rowcolor{gray!25}IDEFICS-Instruct \cite{laurencon2023obelics}&0.8461 &0.4017 & 0.0694& 0.4412& 0.1389&0.0000 &0.3467 & 0.0689 & 0.0896& 0.0528& \bf\textcolor{blue}{0.7787}            \\
InstructBLIP (Flan-T5-XL) \cite{dai2023instructblipgeneralpurposevisionlanguagemodels,chung2024scaling}& 0.9446& 0.4677&0.2849 &0.7683 &0.5711 &0.6705 & 0.6410& 0.0233& 0.0528& 0.0195& 0.8214            \\
\rowcolor{gray!25}InstructBLIP (Vicuna-7B) \cite{dai2023instructblipgeneralpurposevisionlanguagemodels,chiang2023vicuna}&0.2895 & 0.5319& 0.0000&0.3652 & 0.8571& \bf\textcolor{red}{0.9029}&0.3248 & 0.1282& \bf\textcolor{red}{0.3690}& 0.1006& 0.7899            \\
InternLM-XComposer-VL \cite{internlmxcomposer2_5}&0.9336 &0.4157 & 0.2241& 0.7636& 0.6519& 0.6519& 0.5547& 0.1548& 0.1758& 0.1218& 0.8190            \\
\rowcolor{gray!25}LLaVa-v1.5-7B \cite{liu2024improved}& 0.3820& 0.3907& 0.0000& 0.7387& \bf\textcolor{blue}{0.8608}& 0.6694& 0.4925& 0.0724 & 0.1411& 0.0603&0.8311             \\
LLaVa-v1.5-13B \cite{liu2024improved}& 0.9253& 0.2809& 0.2619& 0.7828& 0.7385& 0.5119& 0.5640& 0.0605& 0.1460& 0.0502& 0.8524            \\
\rowcolor{gray!25}LLaVa-v1.6-13B \cite{liu2024llavanext}& \bf\textcolor{blue}{0.9516}& 0.3598& \bf\textcolor{blue}{0.3433}& \bf\textcolor{blue}{0.8106}& 0.6563& 0.4818& 0.6340 & 0.0013& 0.1248& 0.0011& 0.8326            \\
mPLUG-Owl1 \cite{ye2023mplug}& 0.6847& 0.3146&0.2113 &0.6079 & 0.2996& 0.5575& 0.4515&0.0625 & 0.0736& 0.0455& 0.8316             \\
\rowcolor{gray!25}mPLUG-Owl2 \cite{ye2024mplug2}& 0.9417& 0.2579&0.2833 &0.7860 & 0.6459& 0.6402& 0.5283& \bf\textcolor{blue}{0.1617}& 0.1975& \bf\textcolor{blue}{0.1303}&0.8409             \\
mPLUG-Owl3 \cite{ye2024mplug3}& 0.9308& \bf\textcolor{blue}{0.5977}& \bf\textcolor{red}{0.3666}&0.7753 & 0.7972& 0.4716& \bf\textcolor{red}{0.6759}&0.0465 & 0.1469& 0.0387&0.8148             \\
\rowcolor{gray!25}Otter-v1 (MPT-7B) \cite{li2023otter,MosaicML2023Introducing}& 0.7228&0.4530 &0.2397 &0.5075 & 0.5318& 0.6792& 0.5013& 0.0032& 0.0115& 0.0027& 0.8037            \\
Qwen-VL \cite{bai2023qwen}& 0.4638& 0.3767& 0.3333&0.4923 &\bf\textcolor{red}{0.8856} &0.7007 & 0.4945& 0.0897& 0.0917& 0.0652& 0.8179            \\
\rowcolor{gray!25}Shikra (Vicuna-7B) \cite{chen2023shikra,chiang2023vicuna} & \bf\textcolor{red}{0.9638}& 0.2590&0.2424 &0.7588 & 0.4542& 0.4881& 0.4879& 0.0363& 0.0505&0.0267 &0.8511             \\
VisualGLM-6B \cite{glm2024chatglm,wang2023cogvlm}& 0.4012& 0.4784& 0.1406& 0.4587& 0.8435& \bf\textcolor{blue}{0.8636}&0.4605 &0.0057 & 0.0735& 0.0048& \bf\textcolor{red}{0.7609}            \\

\bottomrule
\end{tabular}
\vspace{-0.6cm}
\label{tab:performance}
\end{table*}

\section{Benchmark for MLLMs}
\subsection{Experiment Setup}
To investigate the emotion comprehension capabilities of existing MLLMs, a progressive testing methodology from coarse-grained to fine-grained emotion analysis was designed. Initially, 16 prominent and advanced MLLMs are selected for the evaluation, including two closed-source models, GPT-4o \cite{gpt4o} and Gemini-1.5-Pro \cite{team2024gemini}, and 14 open-source models, such as the LLaVa \cite{liu2024improved,liu2024llavanext} and mPLUG-Owl \cite{ye2023mplug,ye2024mplug2,ye2024mplug3} series. In the coarse-grained emotion comprehension phase, all MLLMs are tasked with identifying the emotion category of each AGPI in the MEMO-Bench dataset. It is worthwhile to additionally note that the true sentiment category is set to the $E_T$ of each AGPI derived from the subjective annotation, rather than the sentiment category of prompts. The comprehension performance is assessed using two metrics: the emotion understanding accuracy and error rate, as similarly defined by Equ.~\ref{lab:gacc} and~\ref{lab:gerr}. All AGPIs that are correctly classified in this phase are then included in the fine-grained emotion perception phase, where the MLLMs are asked to assign an emotion level score to each image. To maintain consistency with the subjectively labeled dataset outlined in Sec.~\ref{sec:sub}, additional prompts are applied to constrain the emotion scores between 0 and 5. For the fine-grained emotion comprehension evaluation, four widely used metrics \cite{vitqa,zhang2023geometry,chen2023no,zhang2024gms,zhang2023simple,min2024perceptual,zhou2022pyramid,zhou2023quality} are employed: Spearman Rank Correlation Coefficient (SRCC), Kendall's Rank Correlation Coefficient (KRCC), Pearson Linear Correlation Coefficient (PLCC), and Root Mean Squared Error (RMSE). SRCC and KRCC are used to assess the monotonicity of the predicted sentiment levels relative to $M O S^{e_d}$ obtained in Sec.~\ref{sec:sub}, while PLCC and RMSE measured the precision of the predictions. The first three metrics range from 0 to 1, with values closer to 1 indicating better performance, whereas RMSE should ideally approach 0.

\subsection{Coarse-grained Emotional Classification}

The understanding accuracy, denoted as $UACC$, and the error rate, $UERR$, for various MLLMs in the coarse-grained sentiment recognition phase are presented in Fig.~\ref{fig:res_mllm} and  Table~\ref{tab:performance}. A combined analysis of these results leads to several important insights: 1) mPLUG-Owl3 achieves the highest emotion classification performance on the MEMO-Bench, reaching an accuracy of 0.6759. However, this score indicates that the performance of current MLLMs in emotion classification tasks remains suboptimal; 2) The MLLMs that achieve the best classification performance for each emotion vary, which reflects the uneven performance and distinct biases of different models in recognizing and classifying emotions; 3) A number of MLLMs achieve accuracy rates exceeding 0.9 for HAP emotion classification, suggesting that existing models exhibit a strong understanding of HAP emotions. In contrast, the recognition accuracy for WOR is notably lower, with WOR frequently misclassified as either SAD or NEU. This indicates a significant gap in the ability of MLLMs to accurately comprehend WOR sentiment; 4) Consistent with the findings in Sec.~\ref{sec:quality} and \ref{sec:gen}, the accuracy of MLLMs' understanding of negative emotions is generally lower than that of positive emotions. This trend suggests that the limitations in AI's computational capacity for processing negative emotions are a widespread challenge across current models.

\subsection{Fine-grained Emotional Comprehension }
All MLLMs are assessed on their ability to comprehend fine-grained emotion degrees using the set of correctly categorized AGPIs, and the results are presented in Table~\ref{tab:performance}. The data in Table~\ref{tab:performance} reveals that, although most MLLMs achieved an accuracy of 0.5 or higher in the emotion categorization task, their performance in fine-grained emotion understanding is notably unsatisfactory. This indicates that while MLLMs can categorize emotions with reasonable accuracy, they fail to capture the more nuanced intensity or degree of human emotions. In summary, MLLMs currently exhibit a limited capacity for understanding human emotions, lacking the ability to empathize with or perceive emotions at a level comparable to human emotional comprehension. This significant finding highlights a critical gap in the emotional intelligence of existing MLLMs and provides valuable insights for guiding the design and training of more advanced emotion-aware models in the future.

%% file: sec/6_con.tex
\section{Conclusion}
The question of whether AI possesses emotions is a critical and ongoing area of research. Addressing this question requires considering at least two key aspects: the emotion generation capabilities of Text-to-Image (T2I) models and the emotion comprehension abilities of Multimodal Large Language Models (MLLMs). To investigate these dimensions, this study introduces the MEMO-Bench dataset, which comprises 7,145 AI-generated portrait images (AGPIs) representing six distinct emotions, produced using 12 state-of-the-art T2I models. To ensure the accuracy and reliability of the labeling data, 29 volunteers are recruited to annotate the AGPIs on three dimensions: image quality, emotional content, and the intensity of the emotion, using a multi-task subjective labeling interface developed on Gradio. The emotion generation capabilities of the T2I models are assessed based on the quality and generation accuracy of the AGPIs. The results indicate that the existing T2I models excel in generating high-quality portraits with positive emotions, but show some limitations in generating negative emotions. Additionally, the MEMO-Bench dataset is utilized to evaluate the emotion comprehension capabilities of MLLMs using a progressive testing approach. The findings reveal that while MLLMs can recognize and categorize emotions to some extent, they are unable to assess the intensity or degree of emotions. Overall, the results suggest that current AI models, both T2I and MLLMs, lack a full understanding of or capacity for emotions, highlighting a significant gap in their emotional intelligence.
